\documentclass[runningheads]{llncs}

\usepackage[mobile]{eccv}

\usepackage{eccvabbrv}

\usepackage{graphicx}
\usepackage{booktabs}

\usepackage[accsupp]{axessibility}  % Improves PDF readability for those with disabilities.

\usepackage[bookmarks=false,pagebackref,breaklinks,colorlinks,citecolor=eccvblue]{hyperref}
\usepackage{orcidlink}

\usepackage{subcaption}
\usepackage{amsmath}
\usepackage{amssymb}
\usepackage{dsfont}
\usepackage{bbding}    % \Checkmark, \XSolid, \CheckmarkBold, \XSolidBold, \XSolidBrush
\usepackage{xcolor}
\definecolor{mydarkgreen}{rgb}{0,.7,0}
\usepackage{url}
\usepackage{multirow}
\usepackage{listings}
\usepackage{tabularx} % control cell length and wrap texts in cell
\usepackage{footnote}
\usepackage{enumitem}

\usepackage[linesnumbered,lined,boxed,ruled]{algorithm2e}
\SetKwInOut{Input}{Input}
\SetKwInOut{Output}{Output}
\SetKwInOut{Initialize}{Initialize}

\usepackage{colortbl}  % \rowcolor{declare-color}, \columncolor{declare-color} (i.e., \rowcolor{blue!20})
\colorlet{mylightblue}{white!90!blue}
\colorlet{mylightgreen}{white!90!green}
\colorlet{mylightgray}{white!90!gray}
\newcommand{\comment}[1]{}

\def\bfa#1{\textbf{\color{black}{#1}}}

\newcommand{\rotate}[1]{\rotatebox[origin=c]{90}{#1}}

\begin{document}

% ---------------------------------------------------------------
% TODO REVIEW: Replace with your title
\title{GKDT: General Keypoint Detection Transformer} 

% TODO REVIEW: If the paper title is too long for the running head, you can set
% an abbreviated paper title here. If not, comment out.
\titlerunning{GKDT: General Keypoint Detection Transformer}

% TODO FINAL: Replace with your author list. 
% Include the authors' OCRID for the camera-ready version, if at all possible.
\author{Changsheng Lu\inst{1} \and
Yuxin Chen\inst{1} \and
Haokun Gui\inst{1} \and
Rong Wang\inst{2} \and
Jie Yang\inst{3} \and
Harry Yang\inst{1} \and
Anton van den Hengel\inst{4} \and
Jiaya Jia\inst{1}
}

% TODO FINAL: Replace with an abbreviated list of authors.
\authorrunning{C. Lu et al.}
% First names are abbreviated in the running head.
% If there are more than two authors, 'et al.' is used.

% TODO FINAL: Replace with your institution list.
\institute{Hong Kong University of Science and Technology, Hong Kong, China\\
\email{changshengluu@gmail.com}, \email{jia@cse.ust.hk} \and
Australian National University, Canberra, Australia \and 
Tencent Inc., China \and
Adelaide University, Adelaide, Australia\\
\email{anton.vandenhengel@adelaide.edu.au}}

\maketitle

\begin{abstract}
With the emergence of various pre-trained vision and language models, computer vision is shifting from narrow-domain to open-domain recognition. The construction of a more powerful yet general keypoint detection (GKD) model to support diverse tasks has become increasingly important in the field. To this end, we firstly present a large-scale unified keypoint dataset called MegaKPT. The dataset is composed of over 1.3 million diverse object instances from twenty-nine existing datasets, and enjoys high-quality unified annotations with keypoint text descriptions. Based on MegaKPT, we develop GKDT, a simple, flexible and powerful DINOv3 based Transformer model for General Keypoint Detection. Our GKDT supports visual prompts, text prompts, or both. To enhance model training, we also propose a suite of useful strategies such as mix-modal prompted training and dynamic importance sampling. By testing over 22 test sets with seen or unseen objects, our single GKDT model shows strong performance and generality in detecting keypoints on broad categories, with most categories over 90\% PCK@0.1 accuracy, offering high practical applicability to real-world problems. The dataset, models, and codes will be released at \url{https://github.com/AlanLuSun/General-Keypoint-Detection}.

\end{abstract}

\section{Introduction}\label{sec:intro}
% highlight the importance of keypoint detection in open-domain scenario (GKD)
% re-iterate the history of kd and related issues (both classic image processing based methods and deep learning based methods) --> thus we need GKD
% briefly introduce existing works on prompt-based keypoint detection, and their limitations (lack large-scale datasets, limited generalization, could use better SSL pre-trained DINOv3, etc.) --> thus we need MegaKPT and GKDT
% introduce our MegaKPT dataset and GKDT model, and their advantages
% during large-scale training, we identify the data imbalance issue rooted in in-the-wild datasets. How we mitigate this issue (dynamic importance sampling)
% report some cheering results 
% summarize our contributions

Keypoint detection is a fundamental task in computer vision that aims to localize a set of keypoints on an image. Compared to object detection, keypoint detection provides more fine-grained part understanding and concise semantic and structural cues about objects, which enables many intriguing applications, such as human and animal pose estimation~\cite{xu2023vitpose++,sun2019deep,cao2019openpose,cheng2020higherhrnet} and robotics \cite{zhou2023clothesnet}.

% action recognition~\cite{wang2019comparative,mathis2018deeplabcut,graving2019deepposekit}, biometric recognition~\cite{kowalski2017deep,sun2013deep}, fine-grained image classification~\cite{zhang2014part,tang2020revisiting}, and robotics \cite{zhou2023clothesnet}. 
% [mouse keypoints detection can help science discovery in biology, \eg, DeepLabCut \cite{mathis2018deeplabcut}; vehicle keypoints detection can help autonomous driving, \eg, CarFusion \cite{reddy2018carfusion}; furniture keypoints detection can help AR/VR applications, \eg, Keypoint-5 \cite{wu2016single}; medical keypoints detection can help disease diagnosis, \eg, Cephalometric \cite{wang2016benchmark}. Many applications!]

\begin{figure*}[!t]
  % \vspace{-0.2cm}
  % \setlength{\abovecaptionskip}{0.1cm}
  % \setlength{\belowcaptionskip}{-0.cm}
  \centering
  \includegraphics[width=\linewidth]{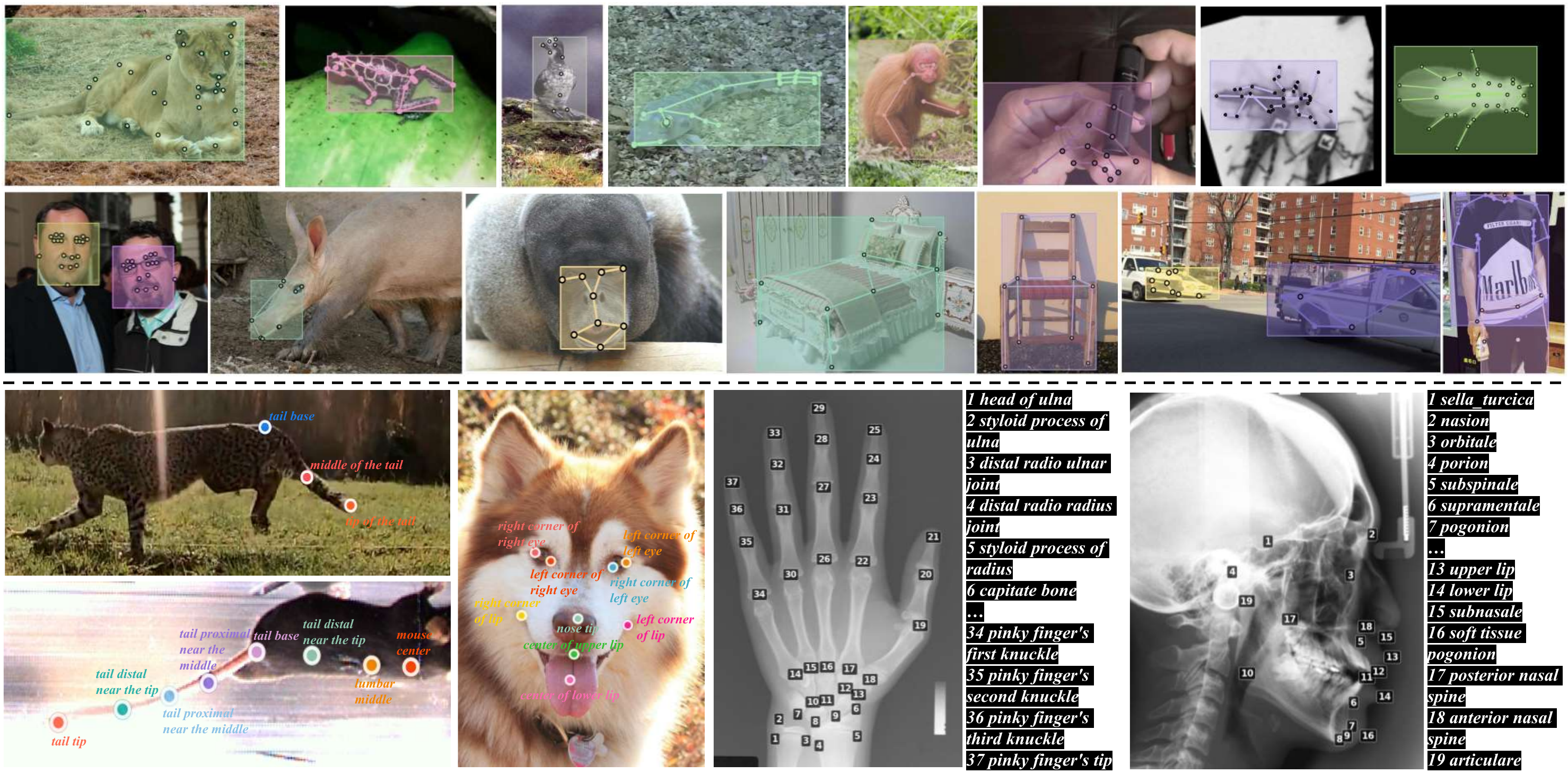}
  \caption{A glance of MegaKPT. All images have both keypoint and text annotations. \textit{(top)} Visualize keypoints only; \textit{(bottom)} Visualize both keypoints and texts. The dataset covers large diversity of objects from various scenes (\textit{best viewed in zoom}).
  %The dataset covers large diversity of objects from natural scene to furniture, vehicle, clothes, and medical images, \etc. (\textit{best viewed when zoomed in})
  % The dataset enjoys large capacity of annotated instances, large object scale variations, and large domain shifts.
  %Our keypoint dataset enjoys the large capacity of annotated object instances, large object scale variations (as small as fly and locust, as large as tiger and elephant), and large domain shift  
  %our large and diverse keypoint dataset. Some example images with keypoint annotations from different super-categories are shown here.
  }
  \label{fig:dataset_examples}
\end{figure*}

Over the past decades, keypoint detection has progressed significantly, from traditional image processing based methods \cite{moravec1980obstacle,harris1988combined,lowe2004distinctive} to deep learning based approaches~\cite{xu2023vitpose++,sun2019deep,cao2019openpose,newell2016stacked,cheng2020higherhrnet}. The traditional methods, \eg, Moravec's detector \cite{moravec1980obstacle} and its successor Harris's detector \cite{harris1988combined}, and SIFT \cite{lowe2004distinctive}, are unsupervised to detect low-level interest points but struggle to localize high-level semantic keypoints. Once we entered deep-learning era, this issue had been overcome, leading to a series of successful convolutional neural network (CNN) or Transformer \cite{vaswani2017attention} based keypoint detection models, expert for various domain-specific categories, such as human pose \cite{cao2019openpose,sun2019deep,xu2023vitpose++} and animal pose models \cite{yu2021ap}. However, these expert models suffer from \emph{close-set detection}, which are unable to recognize keypoints on different kinds of objects, in particular if the objects have quite different anatomies like `fish' \vs `sofa'. 
%\eg, objects that have different numbers of keypoints like `fish' \vs `sofa'. %Moreover, the close-set learning paradigm limits knowledge transferability among domains and model generalization to unseen categories. 
% Moreover, they require large amounts of labeled data for re-training before being able to reconize new class. (\emph{data-inefficient for recognizing novel classes}.)
Therefore, an interesting question emerges: \emph{how to build a strong general keypoint detection (GKD) model practical for diverse keypoints and objects?}

% Why GKD? Open-domain recognition is very important.
% How to detect the object keypoints across diverse species and domains is a fundamental yet challenging problem in computer vision.  
% Existing keypoint detection models are mostly trained in domain-specific datasets, \eg, human pose datasets \cite{lin2014microsoft,ju2023human}, animal pose datasets \cite{yu2021ap,banik2021novel}, face datasets \cite{sagonas2016300,khan2020animalweb}, hand pose datasets \cite{pavlakos2024reconstructing,wang2018mask}, furniture keypoint datasets \cite{wu2016single}, \etc, (Why they only trained in specific individual dataset? Because there is a lack of good unified keypoint dataset!) which limits their generalization to unseen categories and practical applicability. Therefore, building a general keypoint detection (GKD) model that can handle diverse object categories with high accuracy is of great significance.

Inspired by the fact that humans can quickly recognize keypoints given one or a few visual examples, or priors with just language descriptions, recently, researchers have explored prompt based models more general than expert ones, whose goal is to detect keypoints on a query image given one or a few support images with keypoint annotations (\ie, visual prompt) \cite{lu2022few,lu2026exploiting,lu2024detect,xu2022pose,hirschorn2024graph}, keypoint texts (\ie, text prompt) \cite{zhang2023clamp,zhang2024open}, or both \cite{lu2024openkd,yang2024x,lu2024general}, establishing single-modal or multimodal prompted detection. However, most models are still trained in domain-specific or relatively small-scale datasets, which limits the knowledge transferability among domains and model generalization to unseen categories. 

To address this issue, we firstly propose a large-scale unified keypoint dataset, dubbed \textit{\textbf{MegaKPT}}, by unifying 1.3 million diverse object instances from twenty-nine existing datasets into the same annotation format. %, supporting the research of general keypoint detection. 
Compared to the previous dataset UniKPT \cite{yang2024x} which has around 418k instances, our MegaKPT not only has higher volume, but also corrects noisy annotations, supplements accurate keypoint texts and gives clear super-categories and indexes, rendering a high-quality and convenient-to-use dataset. A glance of MegaKPT is shown in Fig.~\ref{fig:dataset_examples}. For the first time, we provide expert keypoint text descriptions for medical scans, such as Cephalometric images for orthodontics \cite{wang2016benchmark} and Hand X-ray images \cite{joham2024implicit}, which makes zero-shot medical landmark detection easy. % medical diagnosis

% we need to give some insights about why we used self-supervised pre-trained visual backbones, such as DINOv3 and MAE ViTs. (unleashing the power of pre-trained visual models for GKD)
To handle the large image diversity of MegaKPT and detect versatile keypoints, we further present \textit{\textbf{GKDT}}, a DINOv3 \cite{simeoni2025dinov3} based Transformer model for General Keypoint Detection. The advent of self-supervised pre-trained vision foundation model (VFM) DINOv3 \cite{simeoni2025dinov3}, could serve as good weight initialization to our visual backbone to extract semantically rich dense features for keypoint localization. Inspired by ViTPose++~\cite{xu2023vitpose++}, we propose a kernel generation (KG) transformer within GKDT to convert both textual and visual keypoint representations into convolution kernels. Afterwards, these convolution kernels will be injected into detection head to perform efficient non-parametric detection. Moreover, to effectively train our model, we propose a suite of useful strategies, including i) mix-modal prompted training to enhance the consistency of prompted patterns between training and test and ii) a novel dynamic importance sampling to mitigate the data imbalance through the perspective of data sampling, which improves scores on tail classes while keeps scores on head ones.

% Despite simplicity, our overall architecture is very flexible and shows strong generality and performance across various seen and unseen categories, with most over 90\% PCK@0.1 accuracy.

% During the large-scale model training, we also encounter the common challenging problem of data imbalance across different super-categories, which is a dilemma of fitting between head and tail classes. If we perform data balance across super-categories, many precious samples will be dropped, weakening the performance of head classes. In this paper, we present a novel sampling strategy called as dynamic importance sampling to mitigate this issue, which greatly improves performance on tail classes while strongly maintain scores on head classes.  

% In particular, our model can zero-shot transfer to medical hand x-ray images with 99.96\% accuracy without any fine-tuning on base keypoints.

Our contributions are in three aspects: \textbf{1)} We elaborately construct a large-scale and diverse keypoint dataset MegaKPT to support the research of general keypoint detection (GKD); \textbf{2)} Our GKDT model enjoys simplicity in architecture design, effectiveness in large-scale model training, and strong generality and performance in detecting keypoints across diverse object categories; and \textbf{3)} We propose a suite of strategies effective for GKDT model training. 

% % \renewcommand{\labelenumi}{\roman{enumi}.}
% % \vspace{-0.1cm}
% % \hspace{-1.0cm}
% % \begin{enumerate}[leftmargin=0.6cm]
% \begin{itemize}
% \item \textbf{Dataset}: We elaborately construct a large-scale and diverse keypoint dataset MegaKPT to support the research of general keypoint detection.
% \item \textbf{Model design}: %Our designed DINOv3 based GKDT model is very flexible and scalable. %, which can incorporate various self-supervised pre-trained visual foundation models, such as DINOv3 and MAE ViTs. 
% Our GKDT model enjoys simplicity in architecture design, effectiveness in large-scale model training, and strong generality and performance in detecting keypoints across diverse object categories.
% \item \textbf{Model training}: We propose a suite of strategies effective for GKD model training, including i) mix-modal prompted training and ii) dynamic importance sampling, to improve multimodal prompted consistency between training and test, and mitigate data imbalance issue, respectively.
% % where the former improves the consistency of multimodal prompted behaviors between training and evaluation, while the latter mitigates the issue of data imbalance and improve model performance, in particular on those tail classes.
% % To mitigate the challenging issue of data imbalance in model training, we propose a simple yet effective data sampling strategy, \ie, dynamic importance sampling, to improve model performance, in particular on those tail classes.
% \end{itemize}
% % \end{enumerate}

To our best knowledge, this work is the first attempt of DINOv3 based model for general keypoint detection. 
% we propose the largest unified keypoint dataset with both visual and text annotations, and 

\section{Related Work}\label{sec:related}
\noindent\textbf{Keypoint Detection} has evolved rapidly from traditional interest point methods~\cite{lowe2004distinctive,harris1988combined} to modern deep learning-based approaches, including deep corner detection~\cite{zhao2023deep}, semi-supervised~\cite{moskvyak2021semi,wang2022pseudo,honari2018improving}, and fully supervised methods~\cite{tompson2014joint,newell2016stacked,cao2019openpose,cheng2020higherhrnet,fang2017rmpe,sun2019deep,xu2023vitpose++}.  
%
% With the rise of convolutional neural networks (CNN)~\cite{krizhevsky2012imagenet,lecun2015deep}, keypoint detection has been significantly advanced by CNN models, \eg, Stacked Hourglass Networks~\cite{newell2016stacked}, OpenPose~\cite{cao2019openpose}, HRNet~\cite{sun2019deep}, and HigherHRNet~\cite{cheng2020higherhrnet}, \etc. More recently, vision transformers (ViT)~\cite{dosovitskiy2020image,vaswani2017attention} have further expanded the design space for keypoint detection. 
%
In general, deep keypoint localization methods can be categorized into two main types: i) direct coordinate regression~\cite{carreira2016human,toshev2014deeppose} and ii) heatmap-based regression with coordinate decoding~\cite{sun2019deep,xu2023vitpose++}. Unlike the existing models that are designed for recognizing specific body parts, such as top-down~\cite{sun2019deep,xu2023vitpose++} or bottom-up~\cite{cao2019openpose,cheng2020higherhrnet} human or animal pose estimators, 
our GKDT model is for general keypoint detection (GKD), which can deal with more versatile keypoints and objects, overcoming the limitations of close-set detection.

% When we entered deep-learning era, keypoint detection has been revolutionized by Convolutional Neural Networks (CNN) \cite{krizhevsky2012imagenet,lecun2015deep}, which led to a series of successful keypoint detection models including the Stacked Hourglass Networks \cite{newell2016stacked}, OpenPose \cite{cao2019openpose}, HRNet \cite{sun2019deep}, HrHRNet \cite{cheng2020higherhrnet}, \etc. In recent years, along with the advent of vision transformer \cite{dosovitskiy2020image,vaswani2017attention} which is a new yet flexible learning architecture, some researchers also explore the transformer based keypoint detection models. The advancements of keypoint detection in the past decade mainly focus on below three aspects: i) better model design, ii) scaling up dataset, \eg, MPII Human Pose \cite{andriluka20142d} and COCO human keypoints \cite{lin2014microsoft}, and iii) efficient optimization, \eg, direct coordinate regression \vs heatmap regression.

\noindent\textbf{General Keypoint Detection} 
% 1. Introduce prompt based models: 
% - single-modal prompted kd (visual prompted, text prompted)
% - multimodal prompted (both)
is a step forward multimodal generic vision aiming at unifying the keypoint detection tasks over various object categories. To this end, the detection model must be very flexible and general. Inspired by few-shot learning \cite{snell2017prototypical,sung2018learning,shi2024few}, a series of works \cite{lu2022few,lu2024detect,lu2026exploiting} %\cite{lu2022few,ge2021metacloth,bohdal2023meta,xu2022pose,hirschorn2024graph} 
explore visual prompted keypoint detection, whose target is to instruct the model to quickly recognize novel or base keypoints on seen or unseen objects given visual prompts. Following this line of research, Chen \etal \cite{chenweak} proposed a new weak-shot setting that transfers keyness and correspondence to novel objects via weakly labeled data.
%Such a setting is quite general as the detection task mainly depends on the input prompts. 
% Xu \etal propose MP-100 dataset by combining 100 category, with each 200 images. Such a setting also inspiring class-agnostic pose estimation works \cite{xu2022pose,hirschorn2024graph}. 

Instead of conditioning on visual prompt, the text prompted keypoint detection \cite{zhang2023clamp,zhang2024open} was proposed thanks to the pre-trained vision-language models (\eg, CLIP~\cite{radford2021learning}) which bridges the gap between vision and language. Compared to visual prompted detection, the text prompted one is more convenient and reconfigurable, saving the efforts to mark keypoints for even one example.

% To embrace multimodal prompting, recently, researchers explore the prompt based model that supports visual prompt, text prompt, or both, enabling a more general keypoint detection. 
To embrace multimodal prompting, some multimodal models were proposed to support both image and text prompting. Lu \etal \cite{lu2024openkd} proposed an OpenKD model, but the OpenKD was trained in small datasets and had limited generality and lower performance on large datasets as a result. %Lu \etal \cite{lu2024openkd} proposed an OpenKD model that opens more prompt diversity, but OpenKD was studied in small-scale datasets and suffered from lower efficiency compared to our GKDT. 
Yang \etal \cite{yang2024x} proposed an end-to-end X-Pose model which predicts object bounding boxes and keypoints simultaneously. However, one model may be hard to regress well for two tasks, and the performance of keypoint detection is restricted by itself object detector. Unlike X-Pose, our GKDT focuses keypoint understanding on individual object, and can couple with more advanced object detectors to deal with multi-object scenarios with higher performance.

% \CL{X-Pose vs. Ours requires more in-depth discussions here. For example, end-to-end vs. two-stage top-down}. Nevertheless, there is still room to further scale up the dataset to train a more general keypoint detection model. Unlike X-pose, we abandons category-level texts for training to avoid inconsistent results and favor keypoint-level texts. In this way, our GKDT model shows stronger part understanding and generality across diverse object categories.

\begin{table}[!tb]
    \centering
    \caption{Statistics of \textbf{MegaKPT}. Each dataset source is credited.
    % Curated datasets used in our simple model training and evaluation. For simplicity, the accumulated counts are reported.
    }
    \label{tab:curated_datasets}
    \resizebox{\linewidth}{!}{  % resize table
    % \scriptsize %\footnotesize %\small
    \fontsize{7}{8}\selectfont
    \setlength{\tabcolsep}{6pt}
    \begin{tabular}{llccrr}
    \toprule[1pt]
    \bfa{Super-category} & \bfa{Dataset} & \bfa{Category}    & \bfa{Keypoint} & \bfa{Image} & \bfa{Instance}\\ \midrule[1pt]
    % \multicolumn{4}{l}{\emph{Super-category: Human pose}} \\
    Human pose  & COCO \cite{lin2014microsoft}               &1            &17         &66,808   &273,469\\ 
                & Human-Art \cite{ju2023human}               &1            &21         &50,000   &123,131\\
    Human face  & 300W \cite{sagonas2016300}                 &1            &68         &600      &600   \\ % combined 4437
                & HELLEN \cite{le2012interactive}            &1            &68         &2,330    &2,330  \\
                & AFW \cite{zhu2012face}                     &1            &68         &337      &337   \\
                & IBUG \cite{sagonas2013300}                 &1            &68         &135      &135   \\
                & LFPW \cite{belhumeur2013localizing}        &1            &68         &1,035    &1,035  \\
                & AFLW \cite{koestinger2011annotated}        &1            &21         &25,993   &25,993 \\
    Human limbs & OneHand10K \cite{wang2018mask}             &1            &21         &11,703   &11,289 \\
                & HInt \cite{pavlakos2024reconstructing}     &1            &21         &17,281   &17,281 \\
    Animal pose & Animal \cite{cao2019cross}                 &5            &20         &4,666    &6,117  \\
                & AwA pose \cite{banik2021novel}             &35           &39         &10,064   &10,064 \\
                & CUB \cite{WahCUB_200_2011}                 &200          &15         &11,788   &11,788 \\
                & NABird \cite{van2015building}              &555          &11         &48,562   &48,562 \\
                & AP-10K \cite{yu2021ap}                     &54           &17         &10,015   &13,028 \\ 
                & APT-36K \cite{yang2022apt}                 &30           &17         &35,708   &48,704 \\
                & MacaquePose \cite{labuguen2021macaquepose} &1            &17         &13,083   &16,393 \\
                & ATRW (tiger) \cite{li2019atrw}             &1            &15         &2,830    &2,830  \\
                & AcinoSet (cheetah) \cite{joska2021acinoset}&1            &20         &5,795    &5,795  \\
                & Animal Kingdom \cite{ng2022animal}         &850          &23         &33,000   &99,267 \\
                & TopViewMouse-5K \cite{ye2022superanimal}   &1            &27         &5,000    &5,000  \\
    Insect pose & Vinegar Fly \cite{pereira2019fast}         &1            &32         &1,500    &1,500  \\
                & Desert Locust \cite{graving2019deepposekit}&1            &35         &700      &700   \\
    Animal face & AnimalWeb \cite{khan2020animalweb}         &350          &9          &22,451   &21,921 \\
    Furniture   & Keypoint-5 \cite{wu2016single}             &5            &8$\sim$14  &8,649    &8,649  \\
    Vehicle     & CarFusion \cite{reddy2018carfusion}        &3            &14         &53,000   &100,000\\
    Clothes     & DeepFashion2 \cite{deepfashion2}           &13           &8$\sim$39  &491,000  &491,000\\
    Medical SC  & Cephalometric \cite{wang2016benchmark}     &1            &19         &400      &400   \\
                & Hand X-ray \cite{joham2024implicit}        &1            &37         &910      &910   \\ \midrule[.5pt]\rowcolor{mylightgreen}
    Accumulated & \textbf{29}                                &\textbf{1587}&\textbf{740}&\textbf{935,343} &\textbf{1,348,228}\\
  % Accumulated & 29                                         &1587 (2118)  &740 (861)  &935,343 &1,348,228\\
    \bottomrule[1pt]
    \end{tabular}
  }
\end{table}

\noindent\textbf{Foundation Models} represent a broad class of models pre-trained on large-scale datasets, including vision foundation models (VFMs) (\eg, DINOv3~\cite{simeoni2025dinov3}), large language models (LLMs), and vision-language models (VLMs), such as CLIP~\cite{radford2021learning}. Using transfer learning technologies \cite{lu2020deep,lu2018viewpoint,wu2021domain} to adapt these models to downstream tasks in an efficient way has become popular in computer vision~\cite{jiao2024toward,liu2023grounding,zhang2024open,zhang2023clamp}. Following this trend, we proactively explore the self-supervised DINOv3 for general keypoint detection.

% \noindent\textbf{Foundation Models} represent a broad class of models pre-trained on large-scale datasets, encompassing vision foundation models (VFMs), such as DINOv3~\cite{simeoni2025dinov3}; large language models (LLMs), including GPT~\cite{achiam2023gpt,brown2020language} and Vicuna~\cite{chiang2023vicuna}; and vision-language models (VLMs), such as CLIP~\cite{radford2021learning}. Recently, transferring the semantically rich knowledge embedded in these models to downstream tasks in an efficient and cost-effective manner has become popular in computer vision~\cite{jiao2024toward,liu2023grounding,zhang2024open,zhang2023clamp}. Following this trend, we proactively explore the \emph{self-supervised} DINOv3 to extract dense image representations for general keypoint localization. Our GKDT model design is very flexible and scalable, allowing seamless integration with various DINOv3 variants to push forward performance limits.

\section{MegaKPT: A Large-Scale High-Quality GKD Dataset}\label{sec:MegaKPT-dataset}
% MegaKPT (ours) vs. UniKPT: Our unified dataset is large and high-quality (clean, low noise, and reliable), but UniKPT is noisy, repetative indexing, unclear super-categories, thus hard to use 
%
% motivation
% summarize the statistics of MegaKPT and comparison
% efforts made to construct MegaKPT
% instance distribution & challenging data imbalance problem

% Along with the progress of general keypoint detection, the pioneering MP-100 dataset \cite{xu2022pose} was proposed to support visual-prompted keypoint detection. MP-100 has 100 categories with each around 200 images. Afterwards, to further increase the capacity of dataset, UniKPT \cite{yang2024x} was proposed. UniKPT has great diversity of categories, however, it suffers from unclear partition of super-categories and repetative categories in annontations, which make it hard to index category and images. (However, since high volume of unified data, UniKPT unavoidably has repetitive category names and ids (hard to index), and unclear partitions of super-categories.  (due to the simply combination of annotation files))

% \CL{Motivation comes first: Why we need to enrich and extend UniKPT, compared to prior works MP-100 and UniKPT? [MP-100 has around 200 instances per category, hard to learn a powerful GKD model pratical to real problems; UniKPT has noisy annotations and repetitive ids, hard to use, cover part of datasets (why not add all), etc.]}

Advancing capacity and diversity of the unified keypoint dataset is crucial for general keypoint detection (GKD). Prior major datasets include MP-100~\cite{xu2022pose} and UniKPT~\cite{yang2024x}. However, MP-100 has around 200 annotated instances per object category, which makes it hard to train a robust GKD model practical for real problems. UniKPT extends MP-100 by incorporating 13 existing datasets, while the annotations are with repetitive category names, noisy indexings, unclear partition of super-categories, and cover partial instances for some datasets. These factors prevent researchers conveniently using it for model training. To address these issues and push forward the unified keypoint dataset, we build a new large-scale high-quality MegaKPT by unifying each dataset from scratch by us. The statistics are shown in Table \ref{tab:curated_datasets}.

% The image sources from twenty-nine datasets are respectively credited. MegaKPT covers over 1.3 million object instances in \emph{unified annotation format} across ten super-categories. All keypoints have text annotations. After filtering the repetitive category names and keypoint texts, UniKPT has accumulated number of 1587 categories and 740 different keypoint types, resulting in great diversity of images.

% To further scale up the diversity of images and support the research of more general keypoint detection, we elaborately prepare a large-scale unified keypoint detection dataset called as MegaKPT, which has over 1.3 million diverse object instances collected from twenty-nine existing public datasets. The statistics of MegaKPT is shown in Table \ref{tab:curated_datasets}. The MegaKPT has 740 unique keypoint types and 1587 object categories from ten super-categories, such as human and animal poses, faces and limbs, insects, furniture, vehicle, clothes, and medical scans. 

\begin{figure}[!tb]
\centering
\begin{minipage}{0.49\textwidth}
  \begin{minipage}{\linewidth}
      \centering
      \setlength{\abovecaptionskip}{0.cm}
      \setlength{\belowcaptionskip}{0.cm}
      \captionof{table}{Statistics of the unified datasets. %Our proposed dataset enjoys significantly larger number of object categories, keypoint types, images, and instances, which can further advance the research of general keypoint detection.
      }
      \label{tab:datasets_comparison}
      % \resizebox{\linewidth}{!}{  % resize table
      \fontsize{6}{6}\selectfont
      \setlength{\tabcolsep}{4.8pt}
      \begin{tabular}{lrrr}
        \toprule[.9pt]
                 & \bfa{MP-100} & \bfa{UniKPT} & \bfa{MegaKPT} \\ \midrule[.9pt]
        Cite     & \cite{xu2022pose}& \cite{yang2024x}  & - \\
        Category & 100       & 1,237          & \bfa{1,587} \\
        Keypoint & -         & 338            & \bfa{740} \\
        Image    & 20,000    & 226,547        & \bfa{935,343} \\
        Instance & 20,000    & 418,487        & \bfa{1,348,228} \\
        \bottomrule[.9pt]
      \end{tabular}
      % }
  \end{minipage}
  \begin{minipage}{\linewidth}
      \centering
      \setlength{\abovecaptionskip}{0.cm}
      \setlength{\belowcaptionskip}{0.cm}
      \captionof{table}{Generalization to unseen hand X-ray images \cite{joham2024implicit} given visual, text and multimodal prompts (\ie, `both'). %Our proposed dataset enjoys significantly larger number of object categories, keypoint types, images, and instances, which can further advance the research of general keypoint detection.
      }
      \label{tab:datasets_generalization}
      % \resizebox{\linewidth}{!}{  % resize table
      \fontsize{6}{6}\selectfont
      \setlength{\tabcolsep}{8.9pt}
      \begin{tabular}{lrrr}
        \toprule[0.9pt]
        PCK@0.1  & \bfa{visual} & \bfa{text} & \bfa{both} \\ \midrule[0.9pt]
        MP-100   & 97.56       & 33.05           & 96.14       \\ 
        UniKPT   & 86.78       & 89.87           & 90.38       \\\rowcolor{mylightgreen} 
        MegaKPT  & \bfa{99.28} & \bfa{99.62}     & \bfa{99.60} \\
        \bottomrule[0.9pt]
      \end{tabular}
      % }
  \end{minipage}
\end{minipage}
\hfill
\begin{minipage}{0.49\textwidth}
  \centering
  % \vspace{-0.2cm}
  \setlength{\abovecaptionskip}{0.cm}
  \setlength{\belowcaptionskip}{-0.1cm}
  \includegraphics[width=\linewidth]{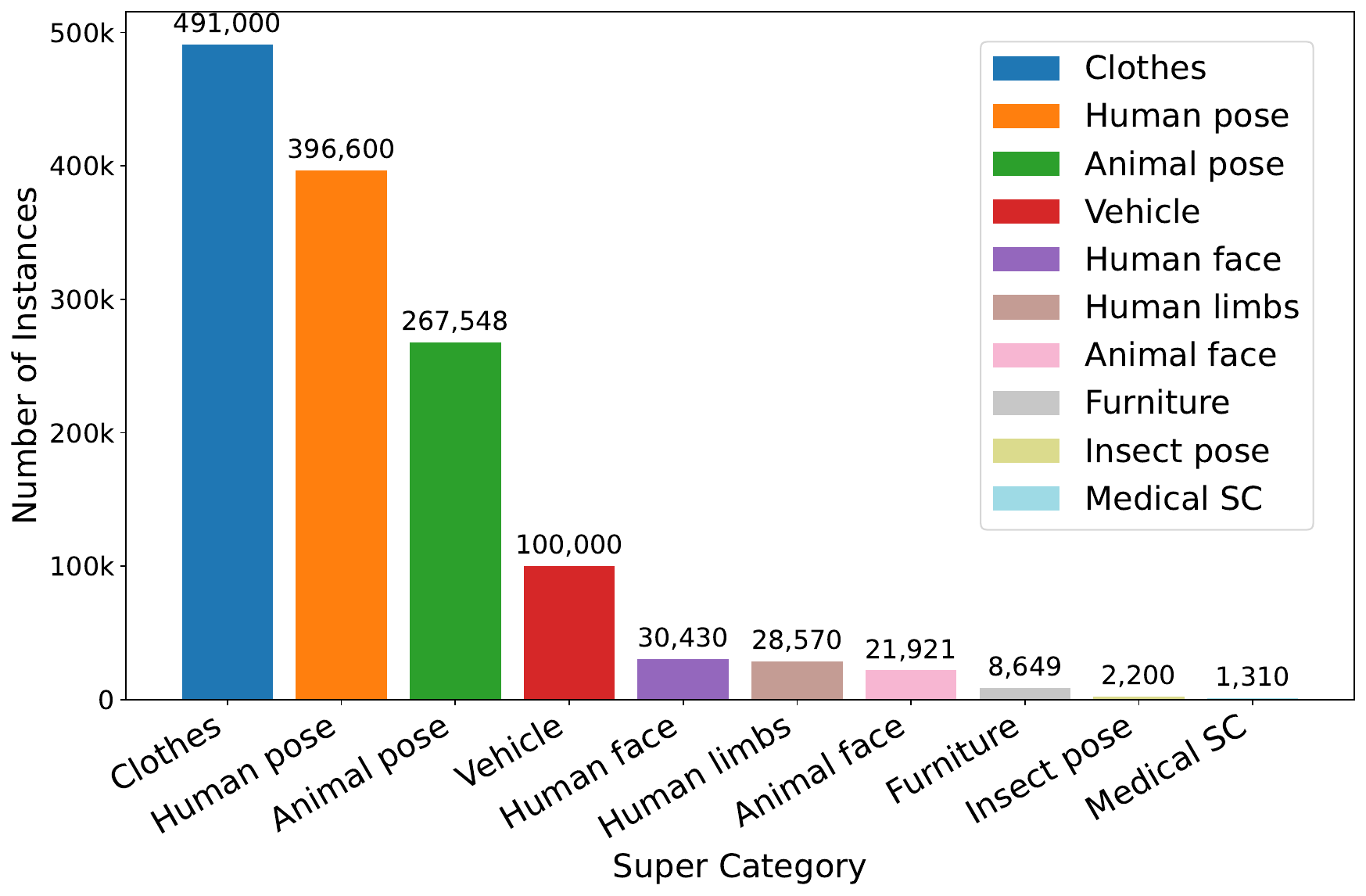}
  \captionof{figure}{Instance distribution against ten super-categories in our unified dataset. %The widely known long-tail phenomenon occurs since \emph{data imbalance} among different super-categories, presenting a challenge for large-scale foundational model training.
  }
  \label{fig:instance_distribution_against_sc}
\end{minipage}
\end{figure}

The proposed MegaKPT has in total over 1.3 million object instances with unified annotation format across ten super-categories, with a distribution shown in Fig.~\ref{fig:instance_distribution_against_sc}. Each keypoint has text annotation. After filtering repetitions, MegaKPT has accumulated number of 740 unique keypoint types and 1587 object categories, spanning human and animal poses, faces and limbs, insects, furniture, vehicle, clothes, and medical scans. As shown in Table~\ref{tab:datasets_comparison}, MegaKPT greatly advances the prior MP-100 and UniKPT. We note that MegaKPT enjoys three features: i) large capacity of data volume; ii) large object scale variations, as small as `vinegar fly' and `locust' while as big as `tiger' and `elephant'; iii) large domain shift, with class ranging from natural scene (\eg, insects and birds) to domestic scene (\eg, furniture and clothes). %(see Fig. \ref{fig:dataset_examples}). %We believe such a great diversity of images would provide a foundation to drive the research of GKD.
These features would well support the GKD research and zero- or few-shot transfer to novel categories.

% \begin{table}[!tb]
%     \centering
%     \caption{Comparison of unified keypoint datasets.
%     }
%     \label{tab:datasets_comparison}
%     % \resizebox{\linewidth}{!}{  % resize table
%     \fontsize{7}{8}\selectfont
%     \setlength{\tabcolsep}{8pt}
%     % \renewcommand{\arraystretch}{1.00}
%     \begin{tabular}{llrrrr}
%     \toprule[1pt]
%     Dataset                   & Venue    &Category  & Keypoints &Images       &Instances \\ \midrule[1pt]
%     MP-100 \cite{xu2022pose}  & ECCV'22  &100       & -         &20,000       &20,000    \\
%     UniKPT \cite{yang2024x}& ECCV'24  &1,237      & 338       &226,547      &418,487   \\
%     MegaKPT (Ours)          & -        &\bfa{1,587}& \bfa{740} &\bfa{935,343}&\bfa{1,348,228} \\
%     \bottomrule[1pt]
%     \end{tabular}
%     % }
% \end{table}
% \begin{figure}[!t]
%   % \vspace{-0.2cm}
%   % \setlength{\abovecaptionskip}{0.1cm}
%   % \setlength{\belowcaptionskip}{-0.cm}
%   \centering
%   \includegraphics[width=.6\linewidth]{pics/supercategory_instances_plot.pdf}
%   \caption{Instance distribution against ten super-categories in \textbf{MegaKPT}. The widely known long-tail phenomenon occurs since \emph{data imbalance} among different super-categories, presenting a challenge for large-scale foundational model training.
%   }
%   \label{fig:instance_distribution_against_sc}
% \end{figure}

%While it may sound easy for curating images from existing datasets, we emphasize that the construction of MegaKPT is non-trivial, as it 
MegaKPT requires significant efforts to manually unify annotations into the same COCO format \cite{lin2014microsoft}, correct ambiguous keypoint locations and texts, and supplement missing keypoint texts for those categories. Moreover, we care the symmetry and relative spatial direction of keypoints, and annotate those keypoint texts in object ego-centric view to ensure uniqueness. We also use LLM \cite{achiam2023gpt} to standardize the object or keypoint names, \eg, `steinbucksteenbok' to `steinbuck steenbok', and `l\_eye' to `left eye', providing a good corpus for keypoint text. Interestingly, for the first time, we give the expert keypoint texts for medical scans of Cephalometric images for orthodontics \cite{wang2016benchmark} and Hand X-ray images \cite{joham2024implicit}, which will benefit zero-shot medical landmark detection (Fig.~\ref{fig:dataset_examples}). Note that all text annotations are manually provided and examined for quality control. 

To explore the dataset advantages, we reserve the entire hand X-ray images for testing while train the same GKDT model in MP-100, UniKPT, and MegaKPT, respectively. All training sets are without hand X-ray images while with real hands in daily life. Table~\ref{tab:datasets_generalization} shows that the model trained in our MegaKPT dataset has highest transfer results across visual (\ie, 1-shot), text (\ie, 0-shot), or both prompting, validating the advancement of MegaKPT. 

% 2. Dataset issue
% MP-100 is a balanced dataset.

% \textbf{Issue of data imbalance:} The instance distribution of our MegaKPT is shown in Fig.~\ref{fig:instance_distribution_against_sc}. As one can see, the head super-categories of clothes, human pose, and animal pose take the dominant number of instances, while the tail super-categories such as medical scans, insect pose, furniture, animal face and human limbs have a total proportion of only 4.65\%, less than 5\%. Such a big data imbalance is an issue rooted in in-the-wild datasets, which presents a challenge for training a general keypoint detection model to not only perform well on head classes but also on tail ones. In Sec. \ref{sec:dynamic_imp_sampling}, we will propose a novel solution through the perspective of data sampling to mitigate this problem.

\begin{figure*}[!t]
  % \vspace{-0.2cm}
  % \setlength{\abovecaptionskip}{0.1cm}
  % \setlength{\belowcaptionskip}{-0.cm}
  \centering
  \includegraphics[width=\linewidth]{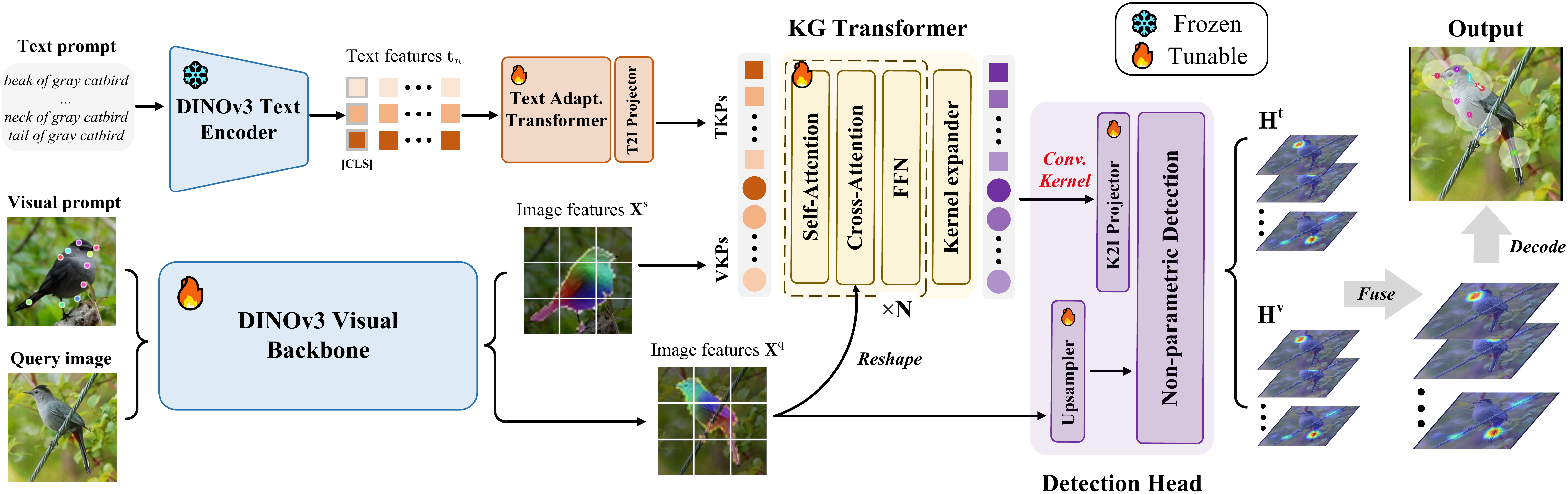}
  \caption{Illustration of the pipeline of our GKDT model for general keypoint detection. Our model takes as input the text prompt, visual prompt, or both to detect the corresponding keypoints on a query image. Our GKDT uses a fully tuned DINOv3 vision transformer as visual backbone to extract semantically rich image features, and a kernel generation (KG) transformer to translate the textual and visual keypoint prototypes (\ie, TKPs \& VKPs) into convolution kernels for efficient non-parametric detection. %Our GKDT can perform strong generality and performance over diverse keypoints on objects.
  % The sketch of model inference. Our OpenKD allows testing under visual prompt, text prompt, or both. For clarification, we show the ``both'' case (\ie, 1-shot with text testing). We firstly extract the deep features of texts, support and query images via CLIP, and then adapt both modalities of features via residual refinement. After extracting the visual keypoint prototype (VKP) and textual counterpart, we build the prototype set to perform class-agnostic correlation and heatmap decoding. Finally, we fuse the heatmaps induced by two modalities (\ie, M1 \& M2) to obtain predictions.
  }
  \label{fig:GKDT_design}
\end{figure*}

\section{General Keypoint Detection Transformer}\label{sec:method}
% \CL{The description of mask token for those invalid kps is missing!}
% \CL{Need to describe the case that if a query keypoint is valid/invalid for a prompt, we will form GT/all-zero heatmap for supervision, which forms positive/negative samples learning. As a result, if a corresponding keypoint does not exist, the predicted heatmap has low heat values.}
% \CL{We use `channel normalize' norm(.) to regularize the energy of conv. kernels to produce stable prompt induced heatmaps. 1/(S*S) * F conv. W}

% setting
Following prior works~\cite{lu2024openkd,yang2024x}, general keypoint detection (GKD) can be trained and evaluated in an episodic manner, where each episode contains a support set and a query set. The query set consists of the images to be detected, while the support set provides the prompts. When \emph{visual prompts} ($K$ support images with marked keypoints) are used, the task is referred to as visual-prompted keypoint detection (\ie, \emph{$K$-shot detection}). When only \emph{text prompt} (language description of keypoint) is given, then the task becomes text-prompted keypoint detection (\ie, \emph{zero-shot} detection). %If \emph{both} given, it is multimodal prompted detection. %The goal of GKD is to detect the corresponding keypoints in the query images based on the provided prompts, whether visual, textual, or both, allowing the model to flexibly handle diverse inputs.

The overall architecture of our GKDT model is shown in Fig.~\ref{fig:GKDT_design}. As one can see, our GKDT is a prompt based model, whose number of output keypoint heatmaps are corresponding to the given input prompts. As such, the model has great flexibility to process the objects that have different numbers of keypoints and anatomy structures, \eg, `tiger', `bird', and `sofa', \etc.

\subsection{Motivation}\label{sec:motivation}
% Overall Architecture
% supports single-modal and multimodal prompted kd

% The overall architecture of our GKDT is shown in Fig.~\ref{fig:GKDT_design}. As one can see, our GKDT is also a prompt based model, whose number of output keypoint heatmaps are corresponding to the given input prompts. As such, the model has great flexibility to process objects that have different number of keypoints and different anatomy structures, \eg, `tiger', `bird', and `sofa'. Such a model flexibility is desired in building general keypoint detection model, which enables the prompt-query feature similarity learning across domains and boosts model generalization. Once the model is trained, it is quite easy to detect the diverse keypoints on seen or unseen species, as the function of the model can be customized by supplying the visual prompt, text prompt, or both. For example, if one would like to detect the `beak', `neck', and `tail' on a bird as shown in Fig.~\ref{fig:GKDT_design}, the detection can be as easy as to instruct the model with the name of these keypoint texts, realizing open-set detection.

The generality of detecting various keypoints is fascinating. However, the question is how to extract representative keypoint prototypes from prompts to effectively guide the model in detecting keypoints in query images, thereby achieving strong performance across diverse objects. 
% We make the hypothesis that the textual and visual keypoint representations would be possible to be refined 
On the one hand, with the advent of vision foundation model (VFM) DINOv3 \cite{simeoni2025dinov3}, it is preferable to choose it as visual backbone to extract semantically dense image features to enhance keypoint representations. As DINOv3 has been self-supervised pre-trained over a large-scale images, its weights can provide a good initialization for model learning. On the other hand, we observe that human pose expert model ViTPose++ \cite{xu2023vitpose++} can achieve state-of-the-art results in COCO benchmark~\cite{lin2014microsoft} even by adding a simple detection head on top of an MAE pre-trained ViT $\mathcal{F}_{\text{v}}$ \cite{he2022masked} as
\begin{equation}
  \mathbf{H} = \text{Conv}_{1\times 1}(\mathcal{U}(\mathcal{F}_{\text{v}}(\mathbf{I}))),
  \label{eq:vitpose}
\end{equation}
where $\mathbf{H} \in \mathbb{R}^{N \times ul \times ul}$ is the $N$ output keypoint heatmaps corresponding to the $N$ predefined keypoints, and $\mathcal{U}$ is an upsampler that can be bilinear interpolation or two deconv blocks to upscale the resolution of image feature map $\mathcal{F}_{\text{v}}(\mathbf{I}) \in \mathbb{R}^{C \times l \times l}$ by ratio $u$. The $\mathbf{I} \in \mathbb{R}^{3 \times l_0 \times l_0}$ is the input image. By taking the convolution weights $\mathbf{W} \in \mathbb{R}^{N \times C \times 1 \times 1}$ out from $\text{Conv}_{1\times 1}$ layer, we can rewrite Eq.~\ref{eq:vitpose} as
\begin{equation}
  \mathbf{H} = \mathcal{U}(\mathcal{F}_{\text{v}}(\mathbf{I})) \otimes \mathbf{W}.
  \label{eq:vitpose2}
\end{equation}
By observing Eq.~\ref{eq:vitpose2}, one can quickly find that each filter $\mathbf{W}_i \in \mathbb{R}^{C \times 1 \times 1}$ acts as similar role to the keypoint prototype to guide the keypoint detection. Inspired by this, if the visual and textual keypoint prototypes (\ie, VKPs \& TKPs) extracted from the prompts could learn keypoint representations well like convolution weights $\mathbf{W}_i$ and exhibit high similarity with image features $\mathcal{F}_{\text{v}}(\mathbf{I})$, our GKDT model has potential to achieve strong performance over open-set object keypoints and categories. This idea directly motivates us to design a kernel generation (KG) transformer inside our GKDT model. In the following, we will present the working mechanism of our GKDT.

\subsection{GKDT Model}
% The core modules of our GKDT include a self-supervised pre-trained DINOv3 visual backbone $\mathcal{F}_{\text{v}}$ \cite{simeoni2025dinov3}, a paired DINOv3 text encoder $\mathcal{F}_{\text{t}}$ \cite{jose2025dinov2}, a text adaptation module $\mathcal{A}_{\text{t}}$, kernel generation (KG) Transformer $\mathcal{K}$, and a detection head $\mathcal{D}_{\text{v}}$.
As shown in Fig.~\ref{fig:GKDT_design}, our GKDT mainly includes four procedures, which are i) image/text feature extraction and adaptation, ii) keypoint prototypes building, iii) kernel generation, and iv) efficient non-parametric detection. 

For brevity, let us assume that the each input episode has one query image $\mathbf{I}^\text{q}$, one support image $\mathbf{I}^\text{s}$ with $N$ annotated keypoints, and $N$ keypoint texts. Firstly, we will use a DINOv3 visual backbone $\mathcal{F}_{\text{v}}$ \cite{simeoni2025dinov3} to encode the support and query images as $\mathcal{F}_{\text{v}}(\mathbf{I}^\text{s})$ and $\mathcal{F}_{\text{v}}(\mathbf{I}^\text{q})$ in deep feature space $\mathbb{R}^{l \times l \times d}$. Moreover, with the DINOv3 text encoder trained by dino.txt $\mathcal{F}_{\text{t}}$ \cite{jose2025dinov2}, the $N$ texts are firstly tokenized, and then encoded as text features $\mathbf{t}_n \in \mathbb{R}^{m \times d}$, where $n=1,2,\cdots, N$; and $m$ is sequence length. For the whole DINOv3 visual backbone, we \emph{fully tune} it to unleash the power of self-supervised learned visual knowledge inside the model for keypoint localization. However, considering the total capacity of keypoint texts are relatively small compared to LLM corpus, we freeze the entire DINOv3 text encoder, and only finetune $\mathbf{t}_n$ with an additional text adaptation transformer $\mathcal{A}_{\text{t}}$ as $\mathbf{t}_n := \mathcal{A}_{\text{t}}(\mathbf{t}_n)$. There is no additional visual adaptation module as the entire DINOv3 visual backbone is tuned. Since only DINOv3 ViT-L has the paired text encoder, to make our model be flexible to accommodate various DINOv3 visual backbone variants, we insert a T2I projector that always maps the text features to have the same channel dimension with the image features, thus handling the feature channel inconsistency.

Afterwards, we extract textual keypoint representations by taking the classification token $\mathbf{\Phi}_{n}^{\text{t}} \in \mathbb{R}^{d}$ from each text feature $\mathbf{t}_n \in \mathbb{R}^{m \times d}$. Then, we perform average to obtain textual keypoint prototype (TKP) as $\mathbf{\Psi}^{\text{t}}_{n}$ if multiple texts are given. For visual keypoint representations (VKR), we encode each support keypoint $\mathbf{p}_n$ into a Gaussian heatmap $\mathbf{H}(\mathbf{p}_n;\sigma)$ and obtain its visual embedding $\mathbf{\Phi}_{n}^{\text{v}} \in \mathbb{R}^{d}$ via linear-weighted summation between $\mathbf{H}(\mathbf{p}_n;\sigma)$ and $\mathcal{F}_{\text{v}}(\mathbf{I}^\text{q})$. Here, $\sigma$ is the standard deviation that controls the spread of the Gaussian. If $K$ support images are provided, the VKRs of the same keypoint type, denoted as $\mathbf{\Phi}_{k,n}^{\text{v}}$, are averaged to produce the visual keypoint prototype (VKP) as $\mathbf{\Psi}^{\text{v}}_{n} = \frac{1}{K}\sum_{k} \mathbf{\Phi}_{k,n}^{\text{v}}$.

Motivated by Section \ref{sec:motivation}, all prototypes are combined as a set of tokens and fed into a KG transformer $\mathcal{K}$ for feature refinement and kernel generation. The $\mathcal{K}$ includes transformers followed by a kernel expander. The self-attention in transformer allows TKPs and VKPs to exchange information and learn relations from each other. In this way, the strong modal prompt, either visual or text prompt, can help the other weak one and improve multimodal prompted performance. % Thus, the visual and text prompts have the opportunity to exchange information. 
To further enhance keypoint representations and improve correlation between the prompt and query image features, we add cross-attention to aggregate more context information from query image. After layer-wise deconv via a kernel expander, the KG will output a set of convolution kernels $\{\mathbf{W}^{\text{v}}_{n}\} \cup \{\mathbf{W}^{\text{t}}_{n}\}=\mathcal{K}(\{\mathbf{\Psi}^{\text{v}}_{n}\} \cup \{\mathbf{\Psi}^{\text{t}}_{n}\}, \mathcal{F}_{\text{v}}(\mathbf{I}^\text{q}))$, where $\mathbf{W}_{n} \in \mathbb{R}^{C \times s \times s}$ is with kernel size $s$. Then, the kernels are projected to the same channel with query image feature via a K2I projector, and performs efficient non-parametric detection as: 
\begin{equation}
  % \begin{aligned}
  \begin{array}{c}
      \mathbf{H}^{\text{v}} = \text{norm}(\mathcal{U}(\mathcal{F}_{\text{v}}(\mathbf{I}^\text{q}))) \otimes \text{norm}(\mathbf{W}^{\text{v}})/s^2 \\
  \mathbf{H}^{\text{t}} = \text{norm}(\mathcal{U}(\mathcal{F}_{\text{t}}(\mathbf{I}^\text{q}))) \otimes \text{norm}(\mathbf{W}^{\text{t}})/s^2
  \end{array},
  % \end{aligned}
\end{equation} 
where $\text{norm}(\cdot)$ refers to $l_2$ normalize in channel dimension. In testing, both the visual or textual prompt induced heatmaps $\mathbf{H}^{\text{v}}$ and $\mathbf{H}^{\text{t}}$ are fused to yield an ensemble heatmap output as $\mathbf{H} = (\mathbb{I}^{\text{v}} \odot \mathbf{H}^{\text{v}} + \mathbb{I}^{\text{t}} \odot \mathbf{H}^{\text{t}}) / (\mathbb{I}^{\text{v}} + \mathbb{I}^{\text{t}})$, 
% \begin{equation}
%   \mathbf{H} = (\mathbb{I}^{\text{v}} \odot \mathbf{H}^{\text{v}} + \mathbb{I}^{\text{t}} \odot \mathbf{H}^{\text{t}}) / (\mathbb{I}^{\text{v}} + \mathbb{I}^{\text{t}}),
% \end{equation}
where $\mathbb{I}^{\text{v}}$ (or $\mathbb{I}^{\text{t}}$) is the validity indicator of keypoints, whose entry $\mathbb{I}^{\text{v}}_{n}$ (or $\mathbb{I}^{\text{t}}_{n}$) is $1$ if the union of $n$-th keypoint in visual prompt (or text prompt) is valid, otherwise is 0. In training, for induced heatmaps, we will construct GT heatmaps $\mathbf{H}^{*}$ for supervision, resulting in the loss $\mathcal{L}=\mathbb{E} \big(\|\mathbb{I}^{\text{v}} \odot (\mathbf{H}^{\text{v}}-\mathbf{H}^{*})\|_\text{F}^2 + \|\mathbb{I}^{\text{t}} \odot (\mathbf{H}^{\text{t}}-\mathbf{H}^{*})\|_\text{F}^2\big) / (\mathbf{1}^{\intercal}\mathbb{I}^{\text{v}} + \mathbf{1}^{\intercal}\mathbb{I}^{\text{t}})$, where $\mathbf{1}$ is all-one vector and $\mathbf{H}^{*}_n=\mathbf{H}(\mathbf{p}^{*}_n;\sigma)$ if the corresponding query keypoint $\mathbf{p}^{*}_n$ exists; otherwise $\mathbf{H}^{*}_n=\mathbf{0}$ which forms negative keypoints learning.

%The learnable modules of our GKDT model are all Transformers and MLP layers, which enjoys simplicity in architecture design. However, our model is quite effective and has strong generality and performance across various categories, which manifests fundamentals speak simply.

% \subsection{Pre-trained Vision Models for GKD}
% % Here we highlight that our model is very flexible to accommodate various visual backbones.
% % Introduce DINOv3 and other pre-trained backbones. 
% % DINOv3, MAE ViT as visual backbone (Unleashing the Power of pre-trained visual models for GKD)(they can provide better visual features to understand image)

% Why we use DINOv3 for GKD? Or we can start by raising a question: Is self-supervised features from DINOv3 or MAE suitable for GKD? 
% Then we try to answer this question... Need to privde in-depth insights here!

% Our GKDT architecture is simple, flexible, and effective, which can replace various backbones to push forward performance. 

\subsection{Strategies for Model Training}\label{sec:strategies_for_training}
% \CL{We propose a suite of strategies effective for GKDT model training: i) mix-modal prompted training (from perspective of prompt sampling) and ii) dynamic importance sampling (from perspective of data sampling)}

To effectively train our model, we propose a suite of strategies as follows: %including i) mix-modal prompted training and ii) dynamic importance sampling.

\textbf{1) Mix-modal prompted training:} When applying the model in real-world scenarios, users may perform visual, text, or both prompting given their preferences. %Thus, the GKD model should support diverse prompts during testing. 
If one only adopts one mode of prompt during training, \eg `both' one, it will inevitably  cause prompt inconsistency between training and test. To mimic real prompted behaviors, we propose mix-modal prompted training from the perspective of prompt sampling, where each training episode randomly selects a mode in set \{\texttt{visual}, \texttt{text}, \texttt{both}\} (default set), and then mask keypoint indicators $\mathbb{I}^{\text{v}}$ and $\mathbb{I}^{\text{t}}$ accordingly. We also studied prompt sets such as \{\texttt{visual}, \texttt{text}\} and \{\texttt{both}\}, and empirically found the default set gives best overall scores.

% \CL{TODO: required to 1) make the text more concise, 2) more vivid (add a figure to show the simulation of sampling process and show distributions are balanced after using dynamic data sampling; see suppl. mat.), 3) add an algorithm, 4) theoretic explanation/discussion (if needed), 5) we can illustrate algorithm by incorporating the math symbols}
% Data distribution imbalance widely occurs in datasets. 
\textbf{2) Dynamic importance sampling:} Fig.~\ref{fig:instance_distribution_against_sc} shows that our MegaKPT exhibits long-tail phenomenon, where the head classes take dominant number of instances while the tail ones take minor. % The tail super-categories including medical scans, insect pose, furniture, animal face and human limbs take a total proportion of only 4.65\%, less than 5\%. 
Such a data imbalance widely exists and presents a challenge for general model training. A common way to address this issue is to perform data balancing. However, it will lose many precious data samples for training and weakens model capability. A nature question is if we could directly train models on such an imbalanced dataset, while not only keep strong performance on head classes but also improve scores on tail ones. 

We try to answer this question through the perspective of data sampling. Consider data sampling without replacement in a data pool $\mathcal{D}=\{D_i\}_{i=1}^S$, where $\mathcal{D}_i$ is the data partition of $i$-th super-category. If one uses uniform sampling, the data samples in tail classes will be quickly exhausted, risking the model in catastrophic forgetting tail classes at late training stage. If using importance sampling, the samples are mostly from head classes, leading to underfitting in tail ones. To reach a balance, we propose a novel dynamic importance sampling, whose key is to dynamically judge if the super-category $c$ of the episode $\mathbf{e}$ sampled via importance $p_c=|\mathcal{D}_c|/\mathcal{D}$ comes from head classes, \ie, those classes with top-$\gamma S$ number of instances, where $\gamma=0.5$ is a fix ratio. If it is, the drawn samples $\mathbf{e}$ will be removed from $\mathcal{D}_c$; otherwise, it will be placed back. If $\mathcal{D}_c$ is empty, the super-category $c$ will be also removed with an update $\mathcal{D}:=\mathcal{D} \backslash \mathcal{D}_c$ and $S := S\!-\!1$. Note that head categories will be dynamically changed based on their remaining number of instances. After iteratively sampling, the whole distribution is gradually balanced. We visualize the distributions in simulated sampling and provide detailed algorithm in \textbf{\S A of Suppl.} %\textbf{\S\ref{sec:appx:dynamic_imp_sampling} of Suppl.} 

% In this way, the number of samples from the head classes will decrease along with the progress of data sampling, and eventually reaching data balance with tail classes. Afterwards, the data samples from the tail classes will be drawn and removed from data pool, too. In this way, in the early stage of data sampling, most of samples are from head classes, but the samples from tail classes have the opportunity to be trained, too. In the late stage, both head and tail classes will have similar opportunity to be added to model training. Such a simple idea directly help us to greatly mitigate data imbalance issue, as evidenced by our experiments.

\begin{table*}[!tb]
  % \vspace{-0.3cm}
  \centering
  \caption{General keypoint detection across thirteen datasets in single-object scenario. The PCK@0.1 score is reported. The test sets that are unseen ($\blacktriangle$) and generalized unseen ($\vartriangle$) are marked. Symbol $^*$ means results reproduced via official released model.
  %One-shot visual prompt is used, and PCK@0.1 scores are reported. All results of our model are obtained via evaluation on \emph{a single model}. %This table shows results on supercategories of animal pose, insect pose and human pose. One can observe that the proposed simple model can achieve generality of keypoint detection across various categories. 
  }
  \label{tab:gkd-benchmark}
  % \small
  \resizebox{\textwidth}{!}{  % resize table
    \scriptsize %\footnotesize %\small
    % \fontsize{7}{8}\selectfont
    \setlength{\tabcolsep}{3pt}
    \begin{tabular}{lcccccccccccccc}
     \toprule[1pt]
      Model &Prompt & \rotate{Animal ($\vartriangle$)}  & \rotate{AwA ($\vartriangle$)}  & \rotate{CUB ($\blacktriangle$)}  & \rotate{NABird ($\blacktriangle$)} & \rotate{AP-10K ($\vartriangle$)} & \rotate{Vinegar fly} & {\rotate{Locust}} & {\rotate{Mouse5k}} & {\rotate{Macaque}} & {\rotate{Tiger}} & {\rotate{Animal Kin.}} & {\rotate{COCO val}} &{\rotate{HumanArt}} \\ \midrule[1pt]%\hline
 DINOv3 \cite{simeoni2025dinov3}&visual &58.24 &60.55 &80.73 &71.11 &59.74 &80.19 &78.10 &50.49 &43.32 &54.17 &40.63 &39.15 &43.34 \\
 CapeF. \cite{shi2023matching}  &visual &41.48 &51.65 &83.91 &87.91 &67.00 &85.36 &80.28 &80.08 &61.32 &60.13 &65.61 &66.78 &62.71 \\
 OpenKD \cite{lu2024openkd}     &visual &49.41 &61.54 &85.51 &86.92	&61.92 &90.87	&90.82 &85.10	&66.60 &69.66	&58.71 &65.26	&60.06 \\\rowcolor{mylightgreen} 
 GKDT (Ours)                    &visual &\bfa{81.11} &\bfa{84.64} &\bfa{97.71} &\bfa{96.25} &\bfa{89.25} &\bfa{97.59} &\bfa{98.66} &\bfa{96.40} &\bfa{88.82} &\bfa{92.33} &\bfa{90.74} &\bfa{88.64} &\bfa{82.92} \\
 \midrule[0.5pt]
 DINOv3 \cite{simeoni2025dinov3}&text   &20.72 &13.82	&10.60 & 8.57 &21.01 & 3.24 & 2.99 & 5.29 &17.45 &22.62 & 5.52 & 3.13 & 2.19 \\
 OpenKD \cite{lu2024openkd}     &text   &55.86 &81.80 &89.82 &91.86 &75.39 &96.95 &97.77 &92.76 &85.40 &87.55 &78.61 &82.40 &77.01 \\
 X-Pose$^*$ \cite{yang2024x}    &text   &40.69 &28.57 &67.07 &39.62 &77.23 &98.13 &98.59 &16.01 &91.60 &55.10 &54.75 &86.05 &73.06 \\\rowcolor{mylightgreen} 
 GKDT (Ours)                    &text   &\bfa{73.27} &\bfa{92.80} &\bfa{98.58} &\bfa{96.94} &\bfa{91.93} &\bfa{98.91} &\bfa{99.49} &\bfa{97.37} &\bfa{93.75} &\bfa{95.88} &\bfa{93.77} &\bfa{93.59} &\bfa{90.48} \\
    \midrule[0.5pt]
 DINOv3 \cite{simeoni2025dinov3}&both   &35.80 &30.77 &44.37 &31.40 &38.64 &23.18 &21.52 &20.61 &33.58 &41.18 &14.91 & 9.36 & 7.36 \\ 
 OpenKD \cite{lu2024openkd}     &both   &57.58 &80.37 &90.82 &91.81 &74.65 &95.97 &96.69 &92.31 &83.89 &86.18 &77.54 &81.35 &75.62 \\\rowcolor{mylightgreen} 
 GKDT (Ours)                    &both   &\bfa{76.87} &\bfa{91.14} &\bfa{98.48} &\bfa{96.86} &\bfa{91.50} &\bfa{98.79} &\bfa{99.37} &\bfa{97.29} &\bfa{93.15} &\bfa{95.40} &\bfa{93.29} &\bfa{93.52} &\bfa{89.68} \\
    \bottomrule[1pt]
    \end{tabular}
 }  % resizebox
% \vspace{-0.3cm}
\end{table*}
% \protect\footnotemark[1]
% \footnotetext[1]{``Generalized unseen'' is usually used to refer that a data sample is from the distributions of seen or unseen categories \cite{xian2017zero}.}

\begin{table*}[!tb]
  % \vspace{-0.3cm}
  \centering
  \caption{General keypoint detection in single-object scenario in nine more datasets. %PCK@0.1 scores are reported.
  % General keypoint detection on nine more datasets (single-object scenario), covering human/animal faces and hands, furniture, vehicles, clothes, and medical scans. i) Results on face and hand datasets. The 300W \cite{sagonas2016300} and AnimalWeb \cite{khan2020animalweb} are human and animal face datasets, respectively. The OneHand10K \cite{wang2018mask} and HInt \cite{pavlakos2024reconstructing} are both human hand datasets. ii) Results on more datasets. We evaluate on furniture (\eg, Keypoint-5 dataset \cite{wu2016single}), vehicles (\eg, CarFusion \cite{reddy2018carfusion}), clothes (\eg, DeepFashion2 dataset \cite{deepfashion2} ), and medical scans (\eg, Cephalometric \cite{wang2016benchmark} and Hand X-ray \cite{joham2024implicit}). We use metric of PCK@0.1 for all scores, and one-shot image for visual prompt. All results are obtained via evaluation on \emph{a single model}. 
  }
  \label{tab:gkd-benchmark-others}
  % \resizebox{\linewidth}{!}{  % resize table
    % \scriptsize %\footnotesize %\small
    \fontsize{5}{5}\selectfont
    \setlength{\tabcolsep}{5.3pt}
    \newcommand{\tabincell}[2]{\begin{tabular}{@{}#1@{}}#2\end{tabular}}
    \begin{tabular}{lcccccccccc}
     \toprule[0.8pt]
\multirow{8}*{Model}&\multirow{8}*{Prompt}& \multicolumn{4}{c}{\bfa{Faces \& hands}} & \multicolumn{3}{c}{\bfa{Furni., vehi., clothes}} & \multicolumn{2}{c}{\bfa{Medical}} \\\cmidrule(lr){3-6}\cmidrule(lr){7-9}\cmidrule(lr){10-11}
            &   & \rotate{300W}  & \rotate{\tabincell{c}{Ani.Web\\($\blacktriangle$)}}  & \rotate{\tabincell{c}{OneHand}}  & \rotate{HInt} &\rotate{Keypoint-5}&\rotate{CarFusion}&\rotate{D.Fashion2}&\rotate{Cephalo}&\rotate{\tabincell{c}{Hand\\X-ray ($\blacktriangle$)}} \\ \midrule[0.8pt]%\hline
 DINOv3 \cite{simeoni2025dinov3}&visual &65.01 &47.37 &29.31 &15.57  &44.89 &32.19  &41.28 &81.67 &91.68\\
 CapeF. \cite{shi2023matching}  &visual &89.47 &71.83 &55.33 &36.38  &53.21 &74.83  &78.76 &67.40 &47.66\\
 OpenKD \cite{lu2024openkd}     &visual &86.98 &64.47 &51.43 &25.66  &67.17 &81.10	&73.94 &97.89	&68.67\\\rowcolor{mylightgreen}  
 GKDT (Ours)                    &visual &\bfa{96.99} &\bfa{82.54} &\bfa{92.56} &\bfa{69.79}  &\bfa{83.01} &\bfa{94.15}  &\bfa{90.96} &\bfa{99.46} &\bfa{99.28}\\ 
 \midrule[0.5pt]
 DINOv3 \cite{simeoni2025dinov3}&text   &16.41 & 8.97 & 7.01 & 4.93  & 2.66 & 0.27  & 1.92 & 0.39 & 1.19\\
 OpenKD \cite{lu2024openkd}     &text   &96.63 &73.34	&75.70 &52.26	 &80.32 &91.85	&87.30 &99.22	&82.24\\
X-Pose$^*$ \cite{yang2024x}     &text   &\bfa{99.53} &27.35	&73.05 &40.42	 &46.02 &87.81	&66.84 & 3.54	& 2.41\\\rowcolor{mylightgreen} 
 GKDT (Ours)                    &text   &99.04 &\bfa{86.36} &\bfa{95.36} &\bfa{82.64}  &\bfa{88.15} &\bfa{96.25}  &\bfa{95.34} &\bfa{99.49} &\bfa{99.62}\\
    \midrule[0.5pt]
 DINOv3 \cite{simeoni2025dinov3}&both   &36.73 &22.11 &14.11 & 9.69  &15.74 & 2.79  & 8.00 &16.08 &25.95\\
 OpenKD \cite{lu2024openkd}     &both   &95.67 &73.10	&73.88 &50.94  &77.87 &90.58	&85.82 &99.18	&85.31\\\rowcolor{mylightgreen}  
 GKDT (Ours)                    &both   &\bfa{98.87} &\bfa{86.24} &\bfa{95.12} &\bfa{81.13}  &\bfa{87.04} &\bfa{95.92}  &\bfa{94.58} &\bfa{99.49} &\bfa{99.60}\\
    \bottomrule[0.8pt]
    \end{tabular}
%  }  % resizebox
% \vspace{-0.3cm}
\end{table*}

\section{Experiments}\label{sec:experiment}
\subsection{Experiment Setup}
\textbf{Dataset Splits:} For fair comparison, we follow original partitions for those datasets in MegaKPT that have official training, validation and test splits, such as COCO \cite{lin2014microsoft} and Human-Art \cite{ju2023human}. Otherwise, we perform splits as follows:
\begin{itemize}
  \item Following previous work \cite{lu2024openkd}, the AwA pose dataset \cite{banik2021novel} is split into 25 species for training and the remaining 10 disjoint ones for testing; The bird datasets CUB \cite{WahCUB_200_2011} and NABird \cite{van2015building} are split as 100, 50, 50 and 333, 111, 111 for training, validation, and test, respectively.
  \item The AP-10K \cite{yu2021ap} uses 42 animal species for training and the remaining 12 species for test. The animal face dataset AnimalWeb \cite{khan2020animalweb} is split into 250 categories as seen classes and the other 100 as unseen ones for test.  
  \item For the purpose of testing model generalization, the entire Animal pose dataset \cite{cao2019cross} and Hand X-ray \cite{joham2024implicit} are reserved for testing.
\end{itemize}
The full splits are in \textbf{\S B of Suppl.}. %\textbf{\S\ref{sec:appx:full_dataset_splits} of Suppl.}. 
Note that test sets of CUB, NABird, AnimalWeb and Hand X-ray are \emph{unseen} to training data. We mark them with ($\blacktriangle$) for distinctions. Moreover, the test sets of Animal pose, AwA and AP-10K are marked with ($\vartriangle$), as they are \emph{generalized unseen}\footnote[1]{`Generalized unseen' refers that a data sample is from seen or unseen classes \cite{xian2017zero}.} with possible overlapping to training data. The training sets are combined for GKD model training.

% In order to fairly compare model performance in existing benchmarks, we ensure that the testing data of each benchmark do not appear in the training phase. To this end, each composed dataset of our unified dataset is split as follows:
% Note that the training data used for our general keypoint detection model is only from the above respective training subsets for fairness.  

\noindent\textbf{Metrics:} Both percentage of correct keypoints (PCK) \cite{lu2022few} and average precision (AP) \cite{lin2014microsoft} (\eg, for COCO and Human-Art) are used for rigorous evaluation. 

%The percentage of correct keypoints (PCK) metric is employed, where a predicted keypoint is considered correct if its distance to the ground truth satisfies $d \leq \rho \cdot \max(w_{\text{bbx}}, h_{\text{bbx}})$, with $w_{\text{bbx}}$ and $h_{\text{bbx}}$ denoting the bounding box width and height, respectively. Following previous works, $\rho$ is set to $0.1$.

\noindent\textbf{Compared Methods:} The compared methods include those that can perform general keypoint detection (GKD) such as the training-free vanilla DINOv3 \cite{simeoni2025dinov3}, OpenKD \cite{lu2024openkd}, and X-Pose \cite{yang2024x}, and the visual prompt based method CapeFormer \cite{shi2023matching} and the methods that are expert for human pose estimation such as state-of-the-art model ViTPose++-H \cite{xu2023vitpose++}. For OpenKD and CapeFormer, we adopt their source codes to fairly train and test in MegaKPT. X-Pose has no training codes as well as visual branches in released model, thus we only compare it for text prompted detection. With different DINOv3 visual backbones, \eg, DINOv3-S/B/L/H, our GKDT has variants GKDT-S/B/L/H accordingly. We use GKDT-L as our default model (\ie, \emph{GKDT}).  

% We will set up two main experiments to demonstrate the effectiveness of our GKDT model. For \emph{general keypoint detection (GKD)}, we select the vinilla DINOv3 \cite{simeoni2025dinov3} and OpenKD \cite{lu2024openkd} for comparison due to their generality in dealing with diverse keypoints. The vinilla DINOv3 is training-free, while for OpenKD we adopt its official source codes to train and fairly compare in MegaKPT. For \emph{expert human pose estimation}, we compare with the state-of-the-art publicly available models such as ViTPose (huge variant) \cite{xu2023vitpose++} and X-Pose \cite{yang2024x}.

% Our GKDT model is compatible with various sizes of visual backbones, including DINOv3-S, DINOv3-B, DINOv3-L, and DINOv3-H. We choose DINOv3-L as our visual backbone by default in consideration of the balance between performance and computational cost.

\noindent\textbf{Implementation Details:} The size of input images to GDKT models is $384 \times 384$. Our KG module $\mathcal{K}$ uses two transformer blocks and text adaptation $\mathcal{A}_{\text{t}}$ uses one. The size of generated kernels $s$ performs well at $1 \times 1$ and $3 \times 3$. All our GKDT models are trained by 20 epochs using 16 H800 GPUs.

\begin{figure*}[!tb]
  % \vspace{-0.2cm}
  \setlength{\abovecaptionskip}{0.2cm}
  \setlength{\belowcaptionskip}{-0.cm}
  \centering
  \includegraphics[width=\linewidth]{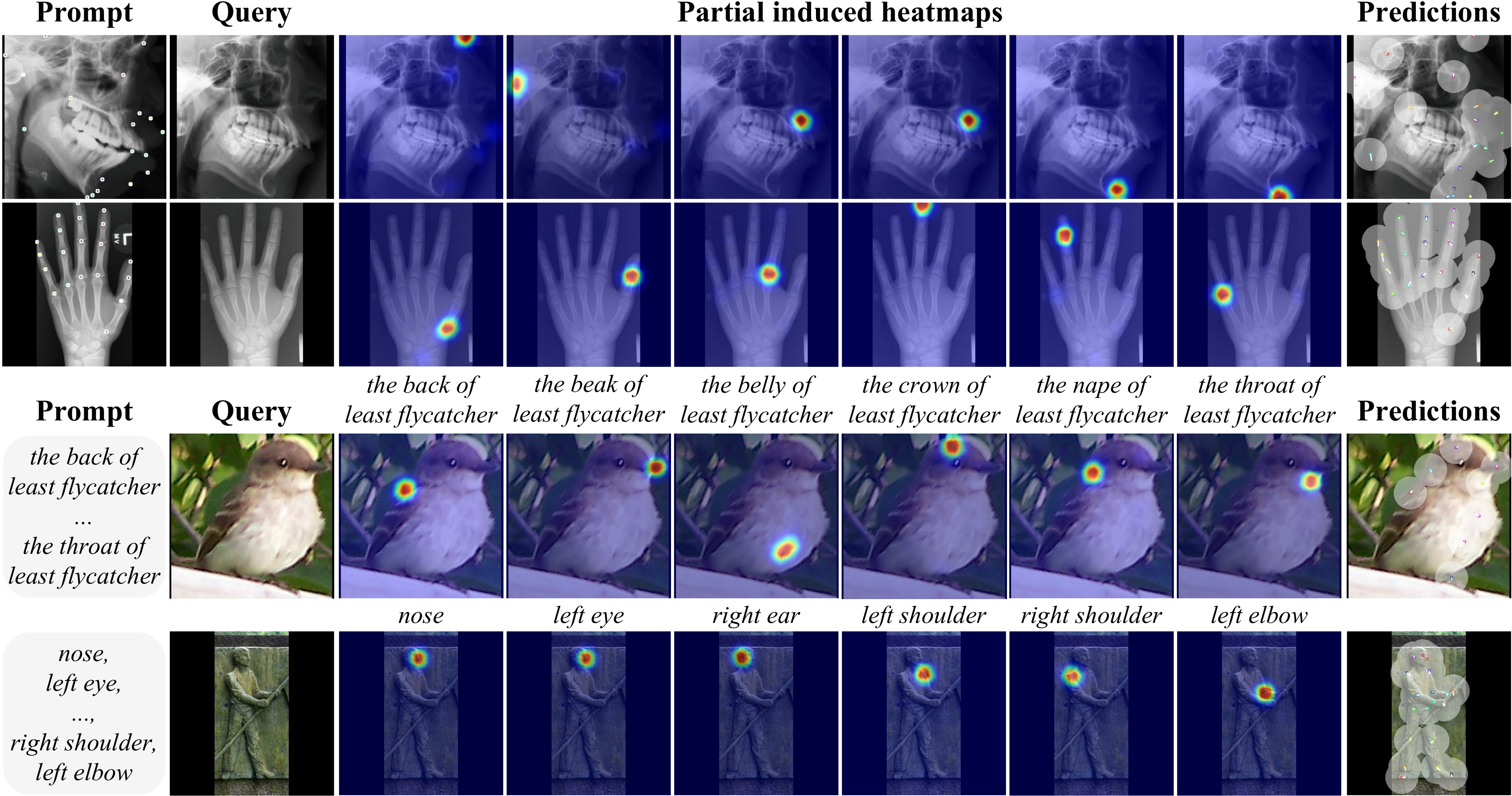}
  \caption{Visualization of visual and text prompted keypoint detection using our GKDT model. The white shadow in predictions signifies PCK@0.1, which means a predicted keypoint (\ie, circle) is correct if falling on this area. GT are marked as tilted crosses.
  %The distance error (short lines) to GT (tilted crosses) are marked as well. % The partial visual and text prompt induced heatmaps are visualized. 
  }
  \label{fig:vis_heatmaps_gkd}
  % \vspace{-0.4cm}
\end{figure*}

\subsection{General Keypoint Detection in Single-Object Scenario}\label{sec:single-object-scenario}

%======PART ONE======
% 1. The training-free vanilla DINOv3 achieves favorable results in visual prompted detection, validating it can provide semantically rich visual features, and serve as a good visual backbone for our GKDT model.
% 2. Table~\ref{tab:gkd-benchmark} shows that our model consistently achieves best results across 13 datasets, whether using visual, text, or both prompts, where 11 out of 13 datasets have over 90\% accuracy under multimodal prompting (\ie, `both' one), showing the generality across various categories.
% We first evaluate models in 13 datasets, spanning three super-categories, which are \emph{animal pose} (\eg, Animal pose dataset \cite{cao2019cross}, AwA pose \cite{banik2021novel}, CUB \cite{WahCUB_200_2011}, NABird \cite{van2015building} datasets, \etc), \emph{insect pose} (\eg, Desert Locust \cite{graving2019deepposekit}, Vinegar Fly \cite{pereira2019fast} datasets), and \emph{human pose} (\eg, COCO \cite{lin2014microsoft}, Human-Art \cite{lin2014microsoft} datasets).

We first perform extensive experiments on general keypoint detection in 22 test sets (Table~\ref{tab:gkd-benchmark} \& Table~\ref{tab:gkd-benchmark-others}), where each input image has a single object cropped via GT bounding box. 1000 episodes with each 10 query images are tested. We use PCK@0.1 as a unified metric for convenient evaluation for all test sets. For COCO and HumanArt, we also report AP \cite{lin2014microsoft} scores in Sec.~\ref{sec:multi-object-scenario}. Table~\ref{tab:gkd-benchmark} shows that our GKDT model consistently achieves best scores and outperforms vanilla DINOv3, OpenKD, and X-Pose under whether visual, text, or both prompts across 13 datasets, covering the super-categories of animal, insect, and human poses. Moreover, we evaluates 9 more datasets covering human and animal faces and hands, furniture, vehicle, clothes, and medical images in Table~\ref{tab:gkd-benchmark-others}. Again, our GKDT shows strong results using a single model. Despite that some test sets are unseen ($\blacktriangle$) or generalized unseen ($\vartriangle$), our model shows good generalization. For instance, our GKDT achieves 98.48\% in CUB and 96.86\% in NABird in localizing the landmarks on unseen birds, 86.24\% in AnimalWeb in unseen animal faces, and 99.60\% for hand X-ray images under multimodal prompting (\ie, `both' one). We suspect that the reason behind the surprising scores on hand X-ray is because medical images are captured in structured scene thus being easier compared to those real hand poses from complex daily scenarios. 
%======PART TWO======
% 1. strong results on 9 more datasets, covering human and animal faces and hands, furniture, vehicle, clothes, and medical datasets in Table~\ref{tab:gkd-benchmark-others}.
%
% 2. our model has strong generalization to unseen or generalized unseen categories. Despite that the tested birds in CUB/NABird, animal faces in AnimalWeb, and Hand X-ray images are unseen during training phase, our GKDT model shows strong generalization/transfer results on them. 
%
% % Hand X-ray dataset \cite{joham2024implicit}
% 3. We evaluate the base keypoints shared with OneHand10K and HInt datasets whose human hand images are collected from daily scenarios. Surprisingly, our GKDT model shows strong zero-shot detection to medical images, achieving over 99\% PCK@0.1 accuracy. We suspect that the high scores are because the hand x-ray images are captured in structured scene thus being easier compared to those hand pose images from complex daily scenarios. Nevertheless, this result demonstrates the promising application of our general keypoint detection model in medical domain.
%
%
%
%======OVERALL======
Together, 17 out of 22 datasets have over 90\% accuracy under multimodal prompting, showing that our GKDT model has strong generality and practical applicability in a broad set of categories. Fig.~\ref{fig:vis_heatmaps_gkd} gives a glance of visualizations.

\begin{table*}[!b]
  % \vspace{-0.3cm}
  \centering
  \caption{Multi-human pose estimation on COCO and Human-Art val sets. General models use text prompts. G-DINO: Grounding DINO \cite{liu2023grounding}; $^*$: results reproduced via official released model; $^\dagger$: finetuned on COCO+HumanArt; l. param: learnable params.  
  % For general models, the zero-shot keypoint detection is performed, \ie, using the text prompt (object class name+kp names). $^*$: results reproduced via official released model; $^\dagger$ means that the model is finetuned on COCO+HumanArt. ``L. Param'' is learnable parameters.
  }
  \label{tab:gkd-benchmark-coco-humanart}
  \scriptsize  % \small 
  \begin{subtable}[b]{.59\textwidth}
    \centering
    \resizebox{\linewidth}{!}{  % resize table
    % \scriptsize %\footnotesize %\small1
    % \fontsize{7}{8}\selectfont
    % \setlength{\tabcolsep}{3pt}
    % \renewcommand{\arraystretch}{1.00}
    % \newcommand{\tabincell}[2]{\begin{tabular}{@{}#1@{}}#2\end{tabular}}
    \begin{tabular}{llcccr}
      \toprule[1pt]
      Models                  & Box Detector   &  AP & AP$^{50}$ & AP$^{75}$ & l.\,param\\ \midrule[1pt]
      \multicolumn{5}{l}{\emph{\quad Expert models}} \\
    % HRNet-W48               & Faster RCNN& 76.3& 90.8 & 82.9 &  63.6M  \\  
      ViTPose-H \cite{xu2023vitpose++}& Faster RCNN& 79.1& 91.6 & 85.7 & 637.2M  \\ \midrule[0.5pt]
    % ViTPose++-H             & Faster RCNN& 79.4& 91.9 & 85.7 & 637.2M  \\ \midrule[0.5pt]
      \multicolumn{5}{l}{\emph{\quad General models}} \\
      X-Pose$^*$ \cite{yang2024x}& End-to-end& 71.8& 88.9 & 78.3 & 128.1M  \\
      GKDT (Ours)           & Faster RCNN    & 73.2& 88.7 & 80.1 & 349.1M  \\
    % GKDT$^\dagger$ (Ours) & Faster RCNN    & 75.3& 89.8 & 82.0 & 349.1M  \\
      GKDT (Ours)           & G-DINO         & 75.0& 90.8 & 82.3 & 349.1M  \\
      GKDT (Ours)           & GT box         & 76.5& 93.7 & 83.9 & 349.1M  \\
    % GKDT$^\dagger$ (Ours) & G-DINO         & 77.3& 92.0 & 84.7 & 349.1M  \\
    % GKDT$^\dagger$ (Ours) & GT box         & 78.4& 94.5 & 85.6 & 349.1M  \\
      GKDT-H (Ours)         & G-DINO         & 77.2& 91.6 & 84.3 & 898.4M  \\
      GKDT-H (Ours)         & GT box         & 78.2& 94.5 & 84.7 & 898.4M  \\
      GKDT-H$^\dagger$ (Ours) & G-DINO       & 78.1& 91.7	& 85.1 & 898.4M  \\
      % GKDT-H$^\dagger$ (Ours) & GT box     & 79.4& 94.6 & 85.8 & 898.4M  \\         
      \bottomrule[1pt]
      \end{tabular}
     }  % resizebox
    \caption{}
    \label{tab:gkd-benchmark-coco}
  \end{subtable}
  % \hfill
  % \hspace{-0.3cm}
  \begin{subtable}[b]{.39\textwidth}
    \centering
    \resizebox{\linewidth}{!}{  % resize table
    \begin{tabular}{llc}
      \toprule[.95pt]
      Models                  & Box Detector&  AP \\ \midrule[.95pt]
      \multicolumn{3}{l}{\emph{\quad Expert models}} \\
      % ViTPose-H               & G-DINO     & 62.1 \\ \midrule[0.5pt]
      ViTPose-H \cite{xu2023vitpose++}& GT         & 67.7 \\ \midrule[0.5pt]
      \multicolumn{3}{l}{\emph{\quad General models}} \\
      X-Pose$^*$ \cite{yang2024x}& End-to-end & 70.7 \\
      GKDT (Ours)                & G-DINO     & 70.0 \\
      GKDT (Ours)                & X-Pose     & 73.4 \\
      GKDT (Ours)                & GT box     & 78.1 \\
    % GKDT$^\dagger$ (Ours)      & G-DINO     & 71.2 \\
    % GKDT$^\dagger$ (Ours)      & X-Pose     & 74.2 \\
    % GKDT$^\dagger$ (Ours)      & GT box     & 79.1 \\
      GKDT-H (Ours)              & G-DINO     & 71.3 \\    
      GKDT-H (Ours)              & GT box     & 79.9 \\    
      GKDT-H$^\dagger$ (Ours)    & G-DINO     & 72.0 \\    
      % GKDT-H$^\dagger$ (Ours)  & GT box     & 80.6 \\ 
      \bottomrule[.95pt]
      \end{tabular}
     }  % resizebox
    \caption{}
    \label{tab:gkd-benchmark-humanart}
  \end{subtable}
  % \vspace{-0.3cm}
\end{table*}

\subsection{General Keypoint Detection in Multi-Object Scenario}\label{sec:multi-object-scenario}
% Careful point: Why not perform end-to-end keypoint detection? because object detection can be more advanced, thus better the keypoint detection performance. (we can add a discussion to discuss the respective advantages of end-to-end vs. two-stage)
%
% 0. Our GKDT focus keypoint detection on individual objects. It can also be applied to multi-object cases coupling with object detectors, such as the open-set detector GroundingDINO \cite{liu2023grounding}, forming the two-stage top-down detection paradigm. Discuss respective advantages between end-to-end (convenient) vs. two-stage (better bounding box, higher performance).
%
% 1.1 GKDT and GKDT-H (w. G-DINO) achieve 75.0\% and 77.2\%, ourperform X-Pose. GKDT-H setting state-of-the-art result in COCO, under the constraint of showing great generality on other diverse categories.
% 1.2 In COCO, Faster RCNN (bbox AP: 56.4) < G-DINO (bbox AP: 65.4) < GT (bbox AP: 1.0), better bboxes, higher performance.
% 1.3 finetuning on COCO+HumanArt, improves results.
% 1.4 In HumanArt, our GKDT and GKDT (w. G-DINO) achieves competative results. G-DINO (bbox AP: 25.0) < X-Pose (bbox AP: 57.9) < GT (bbox AP: 1.0). After using X-Pose's bboxes, improves scores significantly. 
% 1.5 In HumanArt, our GKDT and GKDT (w. G-DINO) have 78.1 and 79.9. Big room to improve scores if with better OD detector.

Our GKDT models focus on keypoint detection for individual objects. However, it can be also applied to multi-object scenario via two-stage top-down method, which firstly localizes object bounding boxes via an object detector and then detects the keypoints for each box. Compared to end-to-end method, two-stage one can couple with more advanced object detectors for high-quality detection.
% \eg, open-set object detector G-DINO \cite{liu2023grounding},

\noindent\textbf{Results on multi-human pose estimation:} Firstly, we evaluate our GKDT models in zero-shot detection in COCO \cite{lin2014microsoft} and HumanArt \cite{ju2023human} validation sets. By coupling with open-set object detector Grounding DINO \cite{liu2023grounding} (G-DINO), our GKDT and GKDT-H yields 75.0\% and 77.2\% in COCO (Table~\ref{tab:gkd-benchmark-coco}), setting the state-of-the-art results in general models. In COCO object detection, Faster RCNN, G-DINO, and GT has AP of 56.4\%, 65.4\%, and 100.0\%. By using better bounding boxs, our GKDT strikes higher results. After finetuning, our GKDT-H improves scores further (78.1\% \vs 77.2\%). %closing the gap to expert model ViTPose-H (78.1\% \vs 79.1\%). 
In HumanArt, our GKDT and GKDT-H with G-DINO obtain scores of 70.0\% and 71.3\%, while with GT boxes have 78.1\% and 79.9\%, showing significant room to improve. The G-DINO performs modest in HumanArt. After using the identical boxes from X-Pose, our GKDT improves the results compared to X-Pose (73.4\% \vs 70.7\%).  
% By digging more, we find G-DINO has object AP of 25.0\% in HumanArt, much lower than X-Pose of 57.9\%. 

% 2.1 eval on more datasets using a single model with PCK under different thresholds $\tau in \{0.05, 0.10, 0.20\}$. The lower, the higher precision. 
% 2.2 Both X-Pose and our GKDT perform well on macaques and cars, while our GKDT strongly outperforms X-Pose in AP-10K and Mouse5K.
% 2.3 visualizations

\begin{figure*}[!tb]
  % \vspace{-0.2cm}
  % \setlength{\abovecaptionskip}{0.1cm}
  % \setlength{\belowcaptionskip}{-0.cm}
  \centering
  \includegraphics[width=\linewidth]{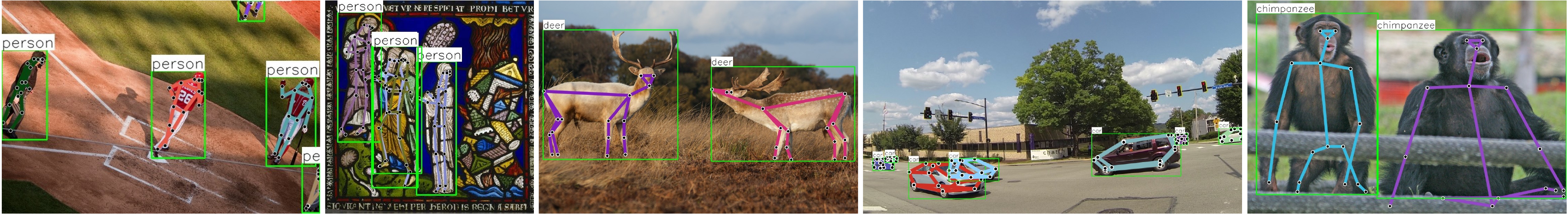}
  \caption{Visualization of general keypoint detection in multi-object scenario via GKDT. %The text prompts are supplied to perform zero-shot detection. The images are from five datasets: COCO, HumanArt, AP-10K, Macaque pose, and Carfusion. Results are from Grounding DINO + GKDT-L.
  }
  \label{fig:vis_multi_object_gkd}
\end{figure*}

\noindent\textbf{Results on animals and vehicles:} We evaluate on more multi-object datasets with PCK under different thresholds. The lower the threshold, the stricter the localization precision. Table~\ref{tab:gkd-benchmark-multi-object} shows that our GKDT model performs quite well, in particular outperforms X-Pose in AP-10K and TopviewMouse5K by a large margin. Note that all the results are achieved via a single model. Fig.~\ref{fig:vis_multi_object_gkd} shows the examples of applying our GKDT in multi-object scenarios.  

% Beyond detecting keypoints on humans, we also evaluate our model on more multi-object scenarios such as animal pose estimation (mouse pose detection can benefit neurobehavior research) and vehicle keypoint detection (can benefit autonomous driving) 
%
% remember to compare the number of learnable model parameters
% Indeed, our model can deal with diverse keypints from different domains, instead of specific domain. The human instances of the Human-Art were collected from quite different domains, ranging from natural scene to cartoon and sculptures, which can be evidenced from their images.

\begin{table*}[!t]
  % \vspace{-0.3cm}
  \centering
  \caption{Results on four multi-object datasets. The scores of PCK@\{0.05, 0.10, 0.20\} are reported. Object boxes are filtered by the score threshold that gives best F-measure. 
  %General keypoint detection (multi-object scenario). We use metric of PCK@\{0.05, 0.10, 0.20\} to showcase the detection precision of compared models.  i) For our method, we first use Grounding DINO to detect the object bounding boxes (by providing the object name as text prompt) and then use GKD to perform detection. ii) X-Pose is end-to-end, which outputs object boxes and keypoints directly. iii) For both methods, the boxes are filtered by using the scores which give best F-measure (There may be many false positive box predictions; F-measure gives good quality for the detected boxes).
  }
  \label{tab:gkd-benchmark-multi-object}
  \resizebox{\linewidth}{!}{  % resize table
    % \scriptsize %\footnotesize %\small
    % \fontsize{7}{8}\selectfont
    \setlength{\tabcolsep}{2pt}
    \newcommand{\tabincell}[2]{\begin{tabular}{@{}#1@{}}#2\end{tabular}}
    \begin{tabular}{llcccccccccccc}
     \toprule[1pt]
\multirow{2}*{Model}&\multirow{2}*{Box Det.}&\multicolumn{3}{c}{\bfa{AP-10K}} & \multicolumn{3}{c}{\bfa{Macaque Pose}} & \multicolumn{3}{c}{\bfa{Mouse5K}} & \multicolumn{3}{c}{\bfa{Carfusion}} \\\cmidrule(lr){3-5}\cmidrule(lr){6-8}\cmidrule(lr){9-11}\cmidrule(lr){12-14}
                            &      & 0.05 & 0.10 & 0.20 & 0.05 & 0.10 & 0.20 & 0.05 & 0.10 & 0.20 & 0.05 & 0.10 & 0.20 \\ \midrule[1pt]%\hline
X-Pose$^*$ \cite{yang2024x} &End-to-end&66.29 &76.06 &82.36 &94.11 &97.41 &98.70 &24.76 &35.43 &44.57 &\bfa{98.96} &\bfa{99.51} &99.78 \\\rowcolor{mylightgreen}   
GKDT (Ours)                 &G-DINO    &\bfa{85.59} &\bfa{92.60} &\bfa{95.56} &\bfa{94.66} &\bfa{97.57} &\bfa{98.78} &\bfa{80.05} &\bfa{82.89} &\bfa{87.37} &98.71 &99.47 &\bfa{99.94} \\    
    \bottomrule[1pt]
    \end{tabular}
 }  % resizebox
% \vspace{-0.3cm}
\end{table*}

\subsection{Ablation Study}
\begin{figure*}[!tb]
  \centering
  \begin{subfigure}[b]{0.31\linewidth}
    \centering
    \includegraphics[width=\linewidth]{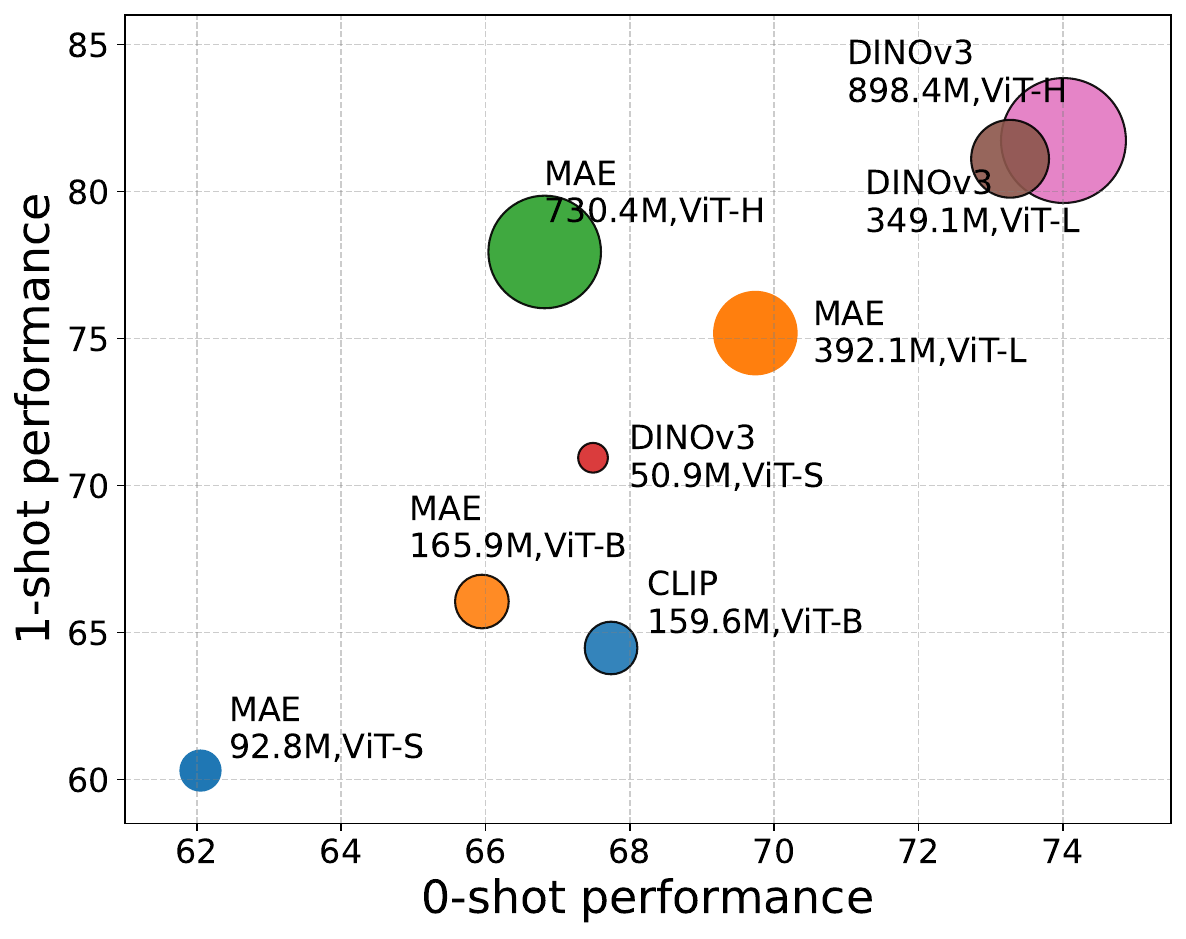}
    \subcaption{}
    \label{fig:study_pretrained_vfms}
    % \hspace{-5pt}
  \end{subfigure}
  \hfill
  \begin{subfigure}[b]{0.36\linewidth}
    \centering
    \includegraphics[width=\linewidth]{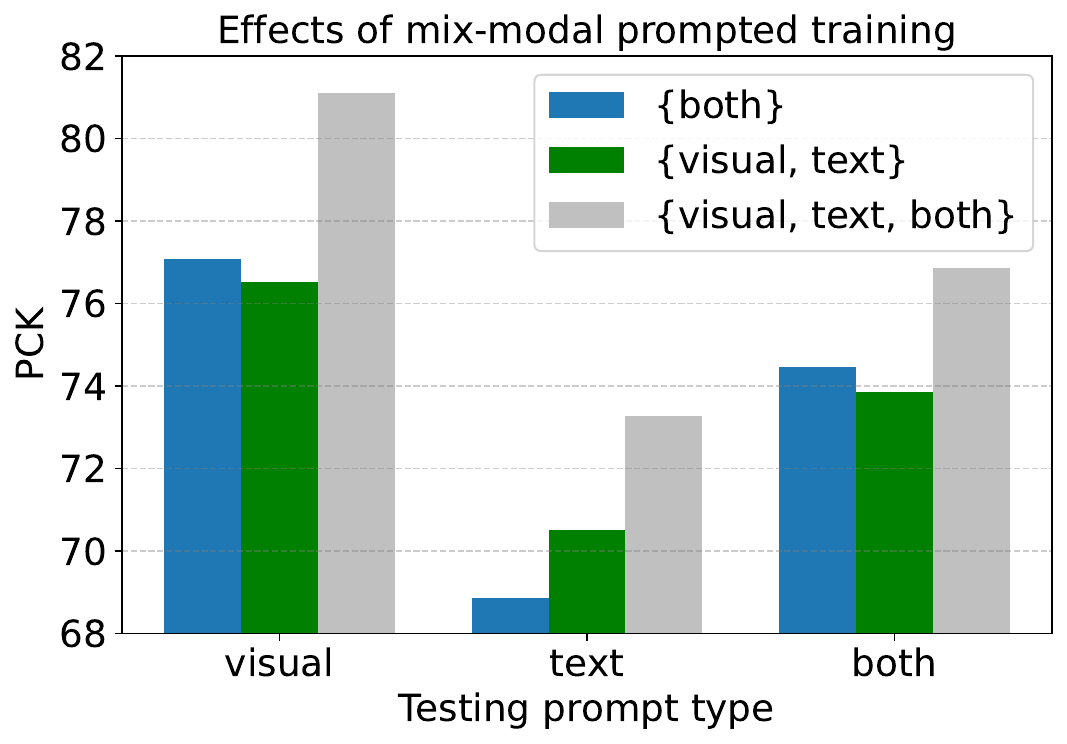}
    \subcaption{}
    \label{fig:study_mix_modal_prompted_training}
  \end{subfigure}
  \hfill
  \begin{subfigure}[b]{0.31\linewidth}
    \centering
    \includegraphics[width=\linewidth]{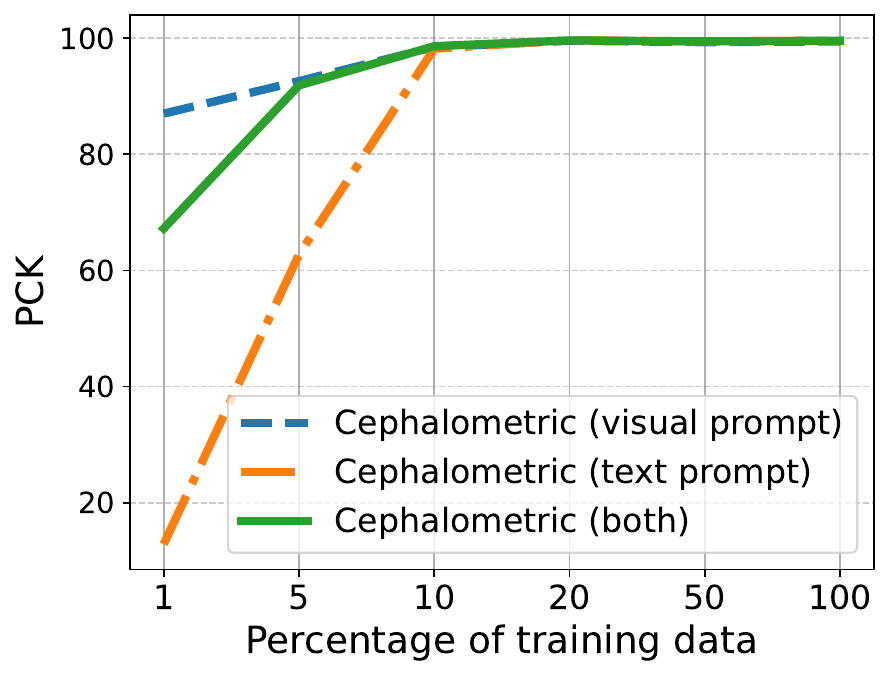}
    \subcaption{}
    \label{fig:study_on_data_amounts_cephalo}
  \end{subfigure}
  \caption{Study on pre-trained VFMs (a), mix-modal prompted training (b), and amounts of training data (c). Animal pose dataset is tested in (a) and (b). PCK@0.1 is used.
  %a) Study on various pre-trained VFMs as visual backbones (VFMs as backbone). The PCK@0.1 scores are reported on \emph{Animal pose dataset}. ; b) Study on impacts of mix-modal prompted training. PCK@0.1 scores are reported in animal pose dataset; c) Study on the influence of training data amounts. PCK@0.1 scores are reported in Cephalometric dataset.
  }
  \label{fig:study_on_misc}
\end{figure*}

\noindent\textbf{Study on pre-trained VFMs:} Besides DINOv3, we investigate using other pre-trained VFMs as visual backbone such as CLIP \cite{brown2020language} and MAE ViTs \cite{he2022masked}. Both DINOv3 and MAE ViTs are self-supervised, while CLIP is weakly-supervised by languages. Fig.~\ref{fig:study_pretrained_vfms} shows that DINOv3 models give best GKD performance and DINOv3-L has performance close to DINOv3-H.

% MAE and DINOv3 are self-supervised pre-trained using images only, while CLIP is weakly-supervised pre-trained by image and language alignment. All pre-trained visual backbones are free of supervision of keypoints. CLIP uses paired visual and text encoder; MAEs uses ViT-B-16; DINOv3 uses DINO.txt. Discovery: 1) DINOv3 provides best visual features overall; 2) the pre-trained text and visual features are not necessarily required to be aligned, as long as the text features could provide good text embeddings. The performance is still good; 3) DINOv3-L has performance very close to DINOv3-H. 

\noindent\textbf{Mix-modal prompted training:} Fig.~\ref{fig:study_mix_modal_prompted_training} shows if only using prompt set \{\texttt{both}\} during training, the single-modal prompted testing, either visual or text, is suboptimal. If using \{\texttt{visual}, \texttt{text}\}, the testing with both prompt drops. By sampling from the set \{\texttt{visual}, \texttt{text}, \texttt{both}\} yields best consistency and results.

\noindent\textbf{Data amounts \& multimodal prompting:} The self-supervised learned visual knowledge and priors of DINOv3 enable strong transfer to GKD. Fig.~\ref{fig:study_on_data_amounts_cephalo} shows that even using 10\% of training data in Cephalometric dataset, our GKDT obtains over 90\% accuracy. Moreover, in real-world testing, one may not know which kind of prompt is optimal. The advantage of using multimodal prompt (\ie, `both' one) can mitigate the risk of choosing a weak-modal prompt and strongly maintain performance compared to the strongest modal prompt, either visual or text ones. In evaluation of using multimodal prompt, Table~\ref{tab:gkd-benchmark} and Table~\ref{tab:gkd-benchmark-others} show that 21 out of 22 tasks have $<2\%$ drops compared to the strongest modal prompt, and 18 out of 22 tasks have $<1\%$ drop.

% Fig.~\ref{fig:study_on_data_amounts_cephalo} shows that even with only 1\% of training data, our GKDT model can achieve over 90\% results on NABird dataset. Moreover, when given multimodal prompts (\ie both visual and text prompts), our model can strongly combine the advantages of either single-modal prompting, namely, visual or text prompting, yielding a good overall performance.

\begin{table*}[!tb]
  % \vspace{-0.3cm}
  \centering
  \caption{Ablations on model components. (a) Impacts of tuning $\mathcal{F}_{\text{v}}$ and using KG transformer $\mathcal{K}$; (b) Context aggregation in $\mathcal{K}$. SA: self-attention; CA: cross-attention.
  %(a) Study on model components (using default model GKDT-L $\mathcal{F}_{\text{v}}$). The $\mathcal{A}_{\text{t}}$ is text transformer; KG is $\mathcal{K}$. (b) Study on the influence of context aggregation in KG ($\mathcal{K}$) [We plan to only use HInt]. SA: self-attention; CA: cross-attention. (aggregate prompted keypoint representations (pose, geometry structure); aggregate context from query image. What are their respective benefits?)  
  }
  \label{tab:study-model-design}
  \scriptsize  % \small 
  \begin{subtable}[t]{.6\textwidth}
    \centering
    \setlength{\tabcolsep}{3pt}
    \begin{tabular}{cccccccc}
     \toprule[1pt]
      \multicolumn{2}{c}{Model}&\multicolumn{2}{c}{Animal pose}  & \multicolumn{2}{c}{CUB} & \multicolumn{2}{c}{HInt} \\\cmidrule(){1-2}\cmidrule(lr){3-4}\cmidrule(lr){5-6}\cmidrule(lr){7-8}
       Tune $\mathcal{F}_{\text{v}}$ &KG                &visual&text &visual&text &visual&text \\ \midrule[1pt]
                                     &                  &53.86 &54.00&79.55 &87.11&12.15 &40.48\\
                                     &\Checkmark        &64.22 &63.88&77.39 &97.23&19.96 &58.57\\
      \Checkmark                     &                  &79.08 &68.29&96.78 &96.81&64.74 &82.83\\\rowcolor{mylightgreen} 
      \Checkmark                     &\Checkmark        &81.11 &73.27&97.71 &98.58&69.79 &82.64\\
    \bottomrule[1pt]
    \end{tabular}
    \caption{}
    \label{tab:study-model-components}
  \end{subtable}
  % \hfill
  % \hspace{-0.3cm}
  \begin{subtable}[t]{.39\textwidth}
    \centering
    \setlength{\tabcolsep}{3.5pt}
    \begin{tabular}{ccclc}
     \toprule[1pt]
      \multicolumn{2}{c}{KG} & \multicolumn{3}{c}{HInt} \\\cmidrule(){1-2}\cmidrule(lr){3-5}
      SA         & CA        &visual&     &text \\ \midrule[1pt]%\hline
                 &           &64.74 &     &82.83\\
      \Checkmark &           &69.05 &\kern-0.5em{(+4.31)}&82.15\\
                 &\Checkmark &65.90 &\kern-0.5em{(+1.16)}&82.94\\\rowcolor{mylightgreen} 
      \Checkmark &\Checkmark &69.79 &\kern-0.5em{(+5.05)}&82.64\\
    \bottomrule[1pt]
    \end{tabular}
    \caption{}
    \label{tab:study-context}
  \end{subtable}
  % \vspace{-0.3cm}
\end{table*}

\noindent\textbf{Study on model components:} Table~\ref{tab:study-model-components} shows that if without tuning visual backbone $\mathcal{F}_{\text{v}}$ and using KG transformer $\mathcal{K}$, the results are worst. After tuning $\mathcal{F}_{\text{v}}$, the scores boost greatly. In both cases of w. and w/o tuning $\mathcal{F}_{\text{v}}$, KG transformer enhances scores and mitigates failures. The self-attention (SA) in $\mathcal{K}$ learns relations and aggregates context among visual and textual keypoint representations, while cross-attention (CA) in $\mathcal{K}$ aggregates context from query image features. To understand their roles, we ablate them. Table \ref{tab:study-context} shows both SA and CA contribute positive impacts, but SA is dominant. Interestingly, the improved scores are in visual prompting, which shows the strong modal prompt (\ie, text one in this case) could improve the weak one.

% Study on the influence of context aggregation in KG ($\mathcal{K}$) [We plan to only use HInt]. SA: self-attention; CA: cross-attention. (aggregate prompted keypoint representations (pose, geometry structure); aggregate context from query image. What are their respective benefits?)

\begin{table*}[!t]
  % \vspace{-0.3cm}
  \centering
  \caption{Study on data sampling strategies and cross-supercategory transfer. (a) Comparison of uniform, importance, and our dynamic importance sampling; (b) Transfer between Human Pose (HP) and Animal Pose (AP) supercategories. S: Source domain; T: Target domain. PCK@0.1 scores under `both' prompt are reported.
  % (a) Comparison of three sampling strategies for model training for general keypoint detection. We perform multimodal prompting and use metric of PCK@0.1 for all scores. The improved scores are in boldface; (b) Study on cross supercategory transfer. S: Source domain; T: Target domain. HP: Human pose supercategory; AP: Animal pose supercategory. Observation: less pose domain shift, higher similarity, thus better performance!
  }
  \label{tab:misc-for-sampling-and-transfer}
  \scriptsize  % \small 
  \begin{subtable}[t]{.69\textwidth}
    \centering
    \setlength{\tabcolsep}{2.5pt}
    \begin{tabular}{llccclcl}
    \toprule[1pt]
\multirow{2}*{{Model}} &\multirow{2}*{{Strategy}}& \multicolumn{2}{c}{{Head}} & \multicolumn{4}{c}{{Tail}} \\\cmidrule(lr){3-4} \cmidrule(l){5-8} 
                     &           &DF2   &COCO  & \multicolumn{2}{c}{HInt}        &\multicolumn{2}{c}{Cephalo}\\\midrule[0.5pt]
\multirow{3}*{OpenKD}&Uniform    &85.23 &81.60 &47.02 &\kern-0.3em{       }&90.99 & \\ 
                     &Importance &85.44 &81.72 &46.64 &\kern-0.3em{       }&92.82 & \\ \rowcolor{mylightgreen} 
                     &Dynamic    &85.82 &81.35 &50.94 &\kern-0.3em{(+3.92)}&99.18 &\kern-0.3em{(+6.36)}\\\midrule[0.5pt]
\multirow{3}*{GKDT}  &Uniform    &94.53 &93.57 &78.18 &\kern-0.3em{       }&99.09 &\\ 
                     &Importance &94.70 &93.40 &80.10 &\kern-0.3em{       }&98.43 &\\ \rowcolor{mylightgreen} 
                     &Dynamic    &94.58 &93.52 &81.13 &\kern-0.3em{(+1.03)}&99.49 &\kern-0.3em{(+0.40)}\\  
    \bottomrule[1pt]
    \end{tabular}
    \caption{}
    \label{tab:dynamic-sampling-part-result}
  \end{subtable}
  % \hfill
  % \hspace{0.2cm}
  \begin{subtable}[t]{.3\textwidth}
    \centering
    \setlength{\tabcolsep}{4.0pt}
    \begin{tabular}{rr}
    \toprule[1pt]
    \multicolumn{1}{c}{{HP to AP}} & {PCK} \\ \midrule[0.5pt]
    COCO (S)   &92.82      \\\rowcolor{mylightgreen}
    Macaque (T)&81.03\\\rowcolor{mylightgreen}
    AP-10K (T) &58.87\\ \midrule[0.5pt]
    \multicolumn{1}{c}{{AP to HP}} & {PCK} \\ \midrule[0.5pt]
    Macaque (S)&90.30      \\
    AP-10K (S) &90.41      \\\rowcolor{mylightgreen}
    COCO (T)   &82.31\\
    \bottomrule[1pt]
    \end{tabular}
    \caption{}
    \label{tab:cross_supercategory_transfer}
  \end{subtable}
  % \vspace{-0.3cm}
\end{table*}

\noindent\textbf{Dynamic importance sampling:} %Data distribution imbalance widely occurs in datasets, which leads a dilema for model training, especially in tail classes. 
Table~\ref{tab:dynamic-sampling-part-result} shows that our dynamic importance sampling greatly enhances model performance in tail classes while keeps scores in head ones, boosting OpenKD by 3.92\% and 6.36\% in multimodal prompting in HInt and Cephalometric datasets, respectively, and in our GKDT by 1.03\% and 0.40\%. We notice that the benefits brought to GKDT are less than OpenKD. We conjecture the reason is as GKDT model is much stronger, thus improvement room is smaller and harder than OpenKD.

% When training a general keypoint detection model on a combined dataset of multiple existing datasets, an obvious issue is the data imbalance. If one performs data balancing, many precious annotated data samples would be dropped, and it may even lead to insufficient learning. However, if one keeps all data samples for training, some categories may be overfitted while others may be underfitted. To mitigate this dilemma, in contrast to using importance sampling for categories, we use dynamic importance sampling to address this issue.

\noindent\textbf{Cross-supercategory transfer:} %By performing transfer experiments between human and animal poses, 
Table~\ref{tab:cross_supercategory_transfer} shows our GKDT model transfers well from human to macaque as they share similar structures (92.82\% \vs 81.03\%), and as expected,  having lower gap compared to four-footed animals in AP-10K. %(92.82\% \vs 58.87\%).

% HP, AP refers to human pose and animal pose. S and T means source and target domains. Interestingly, our GKDT model can transfer well and shows lower gap between human and macaque (92.82\% \vs 81.03\%).
% Observation: less pose domain shift, higher similarity, thus better performance!

\section{Conclusion}\label{sec:conclude}
We advance the unified keypoint dataset to a new level, proposing the MegaKPT to support the research on general keypoint detection. Moreover, our developed DINOv3 based general keypoint detection model, called GKDT, achieves state-of-the-art results in most single-object and multi-object scenarios under whether visual, text, or multimodal prompting, with the help of a suite of useful strategies for model training. We believe that our work will pave the way for GKD research and offers great practicality in real-world applications.

\section*{Acknowledgement}
The authors would like to thank Martin Urschler and Simon Johannes Joham for providing the Hand X-ray dataset, and the helpful suggestions from anonymous reviewers. This work was supported in part by the Research Grants Council under the Areas of Excellence scheme grant AoE/E-601/22-R.

% ---- Bibliography ----
%
% BibTeX users should specify bibliography style 'splncs04'.
% References will then be sorted and formatted in the correct style.
%
\bibliographystyle{splncs04}
\bibliography{refs}

@String(ICCV= {Int. Conf. Comput. Vis.})

@String(AAAI = {AAAI})

@String(ICCV  = {ICCV})

@inproceedings{lu2022few,
  title={Few-shot keypoint detection with uncertainty learning for unseen species},
  author={Lu, Changsheng and Koniusz, Piotr},
  booktitle={Proceedings of the IEEE/CVF conference on computer vision and pattern recognition},
  pages={19416--19426},
  year={2022}
}

@article{lu2026exploiting,
  title={Exploiting Class-agnostic Visual Prior for Few-shot Keypoint Detection},
  author={Lu, Changsheng and Zhu, Hao and Koniusz, Piotr},
  journal={International Journal of Computer Vision},
  volume={134},
  number={2},
  pages={63},
  year={2026},
  publisher={Springer}
}

@inproceedings{lu2024detect,
  title={Detect any keypoints: An efficient light-weight few-shot keypoint detector},
  author={Lu, Changsheng and Koniusz, Piotr},
  booktitle={Proceedings of the AAAI Conference on Artificial Intelligence},
  volume={38},
  number={4},
  pages={3882--3890},
  year={2024}
}

@phdthesis{lu2024general,
  title={General Keypoint Detection: Few-Shot and Zero-Shot},
  author={Lu, Changsheng},
  year={2024},
  school={The Australian National University (Australia)}
}

@inproceedings{xu2022pose,
  title={Pose for everything: Towards category-agnostic pose estimation},
  author={Xu, Lumin and Jin, Sheng and Zeng, Wang and Liu, Wentao and Qian, Chen and Ouyang, Wanli and Luo, Ping and Wang, Xiaogang},
  booktitle={European conference on computer vision},
  pages={398--416},
  year={2022},
  organization={Springer}
}

@inproceedings{shi2023matching,
  title={Matching is not enough: A two-stage framework for category-agnostic pose estimation},
  author={Shi, Min and Huang, Zihao and Ma, Xianzheng and Hu, Xiaowei and Cao, Zhiguo},
  booktitle={Proceedings of the IEEE/CVF Conference on Computer Vision and Pattern Recognition},
  pages={7308--7317},
  year={2023}
}

@inproceedings{hirschorn2024graph,
  title={A Graph-Based Approach for Category-Agnostic Pose Estimation},
  author={Hirschorn, Or and Avidan, Shai},
  booktitle={European Conference on Computer Vision},
  pages={469--485},
  year={2024},
  organization={Springer}
}

@inproceedings{chenweak,
  title={Weak-shot Keypoint Estimation via Keyness and Correspondence Transfer},
  author={Chen, Junjie and Luo, Zeyu and Liu, Zezheng and Jiang, Wenhui and Niu, Li and Fang, Yuming},
  booktitle={The Thirty-ninth Annual Conference on Neural Information Processing Systems},
  year={2025}
}

@inproceedings{zhang2023clamp,
  title={Clamp: Prompt-based contrastive learning for connecting language and animal pose},
  author={Zhang, Xu and Wang, Wen and Chen, Zhe and Xu, Yufei and Zhang, Jing and Tao, Dacheng},
  booktitle={Proceedings of the IEEE/CVF conference on computer vision and pattern recognition},
  pages={23272--23281},
  year={2023}
}

@article{zhang2024open,
  title={Open-vocabulary animal keypoint detection with semantic-feature matching},
  author={Zhang, Hao and Xu, Lumin and Lai, Shenqi and Shao, Wenqi and Zheng, Nanning and Luo, Ping and Qiao, Yu and Zhang, Kaipeng},
  journal={International Journal of Computer Vision},
  volume={132},
  number={12},
  pages={5741--5758},
  year={2024},
  publisher={Springer}
}

@inproceedings{lu2024openkd,
  title={OpenKD: Opening prompt diversity for zero-and few-shot keypoint detection},
  author={Lu, Changsheng and Liu, Zheyuan and Koniusz, Piotr},
  booktitle={European Conference on Computer Vision},
  pages={148--165},
  year={2024},
  organization={Springer}
}

@inproceedings{yang2024x,
  title={X-pose: Detecting any keypoints},
  author={Yang, Jie and Zeng, Ailing and Zhang, Ruimao and Zhang, Lei},
  booktitle={European Conference on Computer Vision},
  pages={249--268},
  year={2024},
  organization={Springer}
}

@article{tompson2014joint,
  title={Joint training of a convolutional network and a graphical model for human pose estimation},
  author={Tompson, Jonathan J and Jain, Arjun and LeCun, Yann and Bregler, Christoph},
  journal={Advances in neural information processing systems},
  volume={27},
  year={2014}
}

@inproceedings{newell2016stacked,
  title={Stacked hourglass networks for human pose estimation},
  author={Newell, Alejandro and Yang, Kaiyu and Deng, Jia},
  booktitle={European conference on computer vision},
  pages={483--499},
  year={2016},
  organization={Springer}
}

@inproceedings{fang2017rmpe,
  title={Rmpe: Regional multi-person pose estimation},
  author={Fang, Hao-Shu and Xie, Shuqin and Tai, Yu-Wing and Lu, Cewu},
  booktitle={Proceedings of the IEEE international conference on computer vision},
  pages={2334--2343},
  year={2017}
}

@inproceedings{sun2019deep,
  title={Deep high-resolution representation learning for human pose estimation},
  author={Sun, Ke and Xiao, Bin and Liu, Dong and Wang, Jingdong},
  booktitle={Proceedings of the IEEE/CVF conference on computer vision and pattern recognition},
  pages={5693--5703},
  year={2019}
}

@article{cao2019openpose,
  title={Openpose: Realtime multi-person 2d pose estimation using part affinity fields},
  author={Cao, Zhe and Hidalgo, Gines and Simon, Tomas and Wei, Shih-En and Sheikh, Yaser},
  journal={IEEE transactions on pattern analysis and machine intelligence},
  volume={43},
  number={1},
  pages={172--186},
  year={2019},
  publisher={IEEE}
}

@inproceedings{cheng2020higherhrnet,
  title={Higherhrnet: Scale-aware representation learning for bottom-up human pose estimation},
  author={Cheng, Bowen and Xiao, Bin and Wang, Jingdong and Shi, Honghui and Huang, Thomas S and Zhang, Lei},
  booktitle={Proceedings of the IEEE/CVF conference on computer vision and pattern recognition},
  pages={5386--5395},
  year={2020}
}

@article{xu2023vitpose++,
  title={Vitpose++: Vision transformer for generic body pose estimation},
  author={Xu, Yufei and Zhang, Jing and Zhang, Qiming and Tao, Dacheng},
  journal={IEEE Transactions on Pattern Analysis and Machine Intelligence},
  volume={46},
  number={2},
  pages={1212--1230},
  year={2023},
  publisher={IEEE}
}

@inproceedings{carreira2016human,
  title={Human pose estimation with iterative error feedback},
  author={Carreira, Joao and Agrawal, Pulkit and Fragkiadaki, Katerina and Malik, Jitendra},
  booktitle={Proceedings of the IEEE conference on computer vision and pattern recognition},
  pages={4733--4742},
  year={2016}
}

@inproceedings{toshev2014deeppose,
  title={Deeppose: Human pose estimation via deep neural networks},
  author={Toshev, Alexander and Szegedy, Christian},
  booktitle={Proceedings of the IEEE conference on computer vision and pattern recognition},
  pages={1653--1660},
  year={2014}
}

@book{moravec1980obstacle,
  title={Obstacle avoidance and navigation in the real world by a seeing robot rover},
  author={Moravec, Hans Peter},
  year={1980},
  publisher={Stanford University}
}

@inproceedings{harris1988combined,
  title={A combined corner and edge detector},
  author={Harris, Chris and Stephens, Mike and others},
  booktitle={Alvey vision conference},
  volume={15},
  number={50},
  pages={10--5244},
  year={1988},
  organization={Manchester, UK}
}

@article{lowe2004distinctive,
  title={Distinctive image features from scale-invariant keypoints},
  author={Lowe, David G},
  journal={International journal of computer vision},
  volume={60},
  number={2},
  pages={91--110},
  year={2004},
  publisher={Springer}
}

@article{zhao2023deep,
  title={Deep Corner},
  author={Zhao, Shanshan and Gong, Mingming and Zhao, Haimei and Zhang, Jing and Tao, Dacheng},
  journal={International Journal of Computer Vision},
  pages={1--25},
  year={2023},
  publisher={Springer}
}

@article{moskvyak2021semi,
  title={Semi-supervised keypoint localization},
  author={Moskvyak, Olga and Maire, Frederic and Dayoub, Feras and Baktashmotlagh, Mahsa},
  journal={arXiv preprint arXiv:2101.07988},
  year={2021}
}

@article{wang2022pseudo,
  title={Pseudo-labeled auto-curriculum learning for semi-supervised keypoint localization},
  author={Wang, Can and Jin, Sheng and Guan, Yingda and Liu, Wentao and Qian, Chen and Luo, Ping and Ouyang, Wanli},
  journal={arXiv preprint arXiv:2201.08613},
  year={2022}
}

@inproceedings{honari2018improving,
  title={Improving landmark localization with semi-supervised learning},
  author={Honari, Sina and Molchanov, Pavlo and Tyree, Stephen and Vincent, Pascal and Pal, Christopher and Kautz, Jan},
  booktitle={Proceedings of the IEEE Conference on Computer Vision and Pattern Recognition},
  pages={1546--1555},
  year={2018}
}

@inproceedings{zhou2023clothesnet,
  title={Clothesnet: An information-rich 3d garment model repository with simulated clothes environment},
  author={Zhou, Bingyang and Zhou, Haoyu and Liang, Tianhai and Yu, Qiaojun and Zhao, Siheng and Zeng, Yuwei and Lv, Jun and Luo, Siyuan and Wang, Qiancai and Yu, Xinyuan and others},
  booktitle={Proceedings of the IEEE/CVF International Conference on Computer Vision},
  pages={20428--20438},
  year={2023}
}

@article{vaswani2017attention,
  title={Attention is all you need},
  author={Vaswani, Ashish and Shazeer, Noam and Parmar, Niki and Uszkoreit, Jakob and Jones, Llion and Gomez, Aidan N and Kaiser, {\L}ukasz and Polosukhin, Illia},
  journal={Advances in neural information processing systems},
  volume={30},
  year={2017}
}

@article{snell2017prototypical,
  title={Prototypical networks for few-shot learning},
  author={Snell, Jake and Swersky, Kevin and Zemel, Richard S},
  journal={arXiv preprint arXiv:1703.05175},
  year={2017}
}

@inproceedings{sung2018learning,
  title={Learning to compare: Relation network for few-shot learning},
  author={Sung, Flood and Yang, Yongxin and Zhang, Li and Xiang, Tao and Torr, Philip HS and Hospedales, Timothy M},
  booktitle={Proceedings of the IEEE/CVF Conference on Computer Vision and Pattern Recognition},
  pages={1199--1208},
  year={2018}
}

@inproceedings{shi2024few,
  title={Few-shot shape recognition by learning deep shape-aware features},
  author={Shi, Wenlong and Lu, Changsheng and Shao, Ming and Zhang, Yinjie and Xia, Siyu and Koniusz, Piotr},
  booktitle={Proceedings of the IEEE/CVF Winter Conference on Applications of Computer Vision},
  pages={1848--1859},
  year={2024}
}

@inproceedings{xian2017zero,
  title={Zero-shot learning-the good, the bad and the ugly},
  author={Xian, Yongqin and Schiele, Bernt and Akata, Zeynep},
  booktitle={Proceedings of the IEEE conference on computer vision and pattern recognition},
  pages={4582--4591},
  year={2017}
}

@article{lu2020deep,
  title={Deep transfer neural network using hybrid representations of domain discrepancy},
  author={Lu, Changsheng and Gu, Chaochen and Wu, Kaijie and Xia, Siyu and Wang, Haotian and Guan, Xinping},
  journal={Neurocomputing},
  volume={409},
  pages={60--73},
  year={2020},
  publisher={Elsevier}
}

@inproceedings{lu2018viewpoint,
  title={Viewpoint estimation for workpieces with deep transfer learning from cold to hot},
  author={Lu, Changsheng and Wang, Haotian and Gu, Chaochen and Wu, Kaijie and Guan, Xinping},
  booktitle={International Conference on Neural Information Processing},
  pages={21--32},
  year={2018},
  organization={Springer}
}

@inproceedings{wu2021domain,
  title={Domain adaptation for viewpoint estimation with image generation},
  author={Wu, Xunjin and Lu, Changsheng and Gu, Chaochen and Wu, Kaijie and Zhu, Shanying},
  booktitle={2021 International Conference on control, automation and information sciences (ICCAIS)},
  pages={341--346},
  year={2021},
  organization={IEEE}
}

@inproceedings{radford2021learning,
  title={Learning transferable visual models from natural language supervision},
  author={Radford, Alec and Kim, Jong Wook and Hallacy, Chris and Ramesh, Aditya and Goh, Gabriel and Agarwal, Sandhini and Sastry, Girish and Askell, Amanda and Mishkin, Pamela and Clark, Jack and others},
  booktitle={International conference on machine learning},
  pages={8748--8763},
  year={2021},
  organization={PMLR}
}

@article{liu2023grounding,
  title={Grounding dino: Marrying dino with grounded pre-training for open-set object detection},
  author={Liu, Shilong and Zeng, Zhaoyang and Ren, Tianhe and Li, Feng and Zhang, Hao and Yang, Jie and Li, Chunyuan and Yang, Jianwei and Su, Hang and Zhu, Jun and others},
  journal={arXiv preprint arXiv:2303.05499},
  year={2023}
}

@article{jiao2024toward,
  title={Toward Re-Identifying Any Animal},
  author={Jiao, Bingliang and Liu, Lingqiao and Gao, Liying and Wu, Ruiqi and Lin, Guosheng and Wang, Peng and Zhang, Yanning},
  journal={Advances in Neural Information Processing Systems},
  volume={36},
  year={2024}
}

@article{simeoni2025dinov3,
  title={Dinov3},
  author={Sim{\'e}oni, Oriane and Vo, Huy V and Seitzer, Maximilian and Baldassarre, Federico and Oquab, Maxime and Jose, Cijo and Khalidov, Vasil and Szafraniec, Marc and Yi, Seungeun and Ramamonjisoa, Micha{\"e}l and others},
  journal={arXiv preprint arXiv:2508.10104},
  year={2025}
}

@inproceedings{jose2025dinov2,
  title={Dinov2 meets text: A unified framework for image-and pixel-level vision-language alignment},
  author={Jose, Cijo and Moutakanni, Th{\'e}o and Kang, Dahyun and Baldassarre, Federico and Darcet, Timoth{\'e}e and Xu, Hu and Li, Daniel and Szafraniec, Marc and Ramamonjisoa, Micha{\"e}l and Oquab, Maxime and others},
  booktitle={Proceedings of the Computer Vision and Pattern Recognition Conference},
  pages={24905--24916},
  year={2025}
}

@inproceedings{he2022masked,
  title={Masked autoencoders are scalable vision learners},
  author={He, Kaiming and Chen, Xinlei and Xie, Saining and Li, Yanghao and Doll{\'a}r, Piotr and Girshick, Ross},
  booktitle={Proceedings of the IEEE/CVF conference on computer vision and pattern recognition},
  pages={16000--16009},
  year={2022}
}

@article{brown2020language,
  title={Language models are few-shot learners},
  author={Brown, Tom and Mann, Benjamin and Ryder, Nick and Subbiah, Melanie and Kaplan, Jared D and Dhariwal, Prafulla and Neelakantan, Arvind and Shyam, Pranav and Sastry, Girish and Askell, Amanda and others},
  journal={Advances in neural information processing systems},
  volume={33},
  pages={1877--1901},
  year={2020}
}

@article{achiam2023gpt,
  title={GPT-4 Technical Report},
  author={Achiam, Josh and Adler, Steven and Agarwal, Sandhini and Ahmad, Lama and Akkaya, Ilge and Aleman, Florencia Leoni and Almeida, Diogo and Altenschmidt, Janko and Altman, Sam and Anadkat, Shyamal and others},
  journal={arXiv preprint arXiv:2303.08774},
  year={2023}
}

@inproceedings{lin2014microsoft,
  title={Microsoft coco: Common objects in context},
  author={Lin, Tsung-Yi and Maire, Michael and Belongie, Serge and Hays, James and Perona, Pietro and Ramanan, Deva and Doll{\'a}r, Piotr and Zitnick, C Lawrence},
  booktitle={European conference on computer vision},
  pages={740--755},
  year={2014},
  organization={Springer}
}

@inproceedings{ju2023human,
  title={Human-art: A versatile human-centric dataset bridging natural and artificial scenes},
  author={Ju, Xuan and Zeng, Ailing and Wang, Jianan and Xu, Qiang and Zhang, Lei},
  booktitle={Proceedings of the IEEE/CVF conference on computer vision and pattern recognition},
  pages={618--629},
  year={2023}
}

@article{sagonas2016300,
  title={300 faces in-the-wild challenge: Database and results},
  author={Sagonas, Christos and Antonakos, Epameinondas and Tzimiropoulos, Georgios and Zafeiriou, Stefanos and Pantic, Maja},
  journal={Image and vision computing},
  volume={47},
  pages={3--18},
  year={2016},
  publisher={Elsevier}
}

@inproceedings{le2012interactive,
  title={Interactive facial feature localization},
  author={Le, Vuong and Brandt, Jonathan and Lin, Zhe and Bourdev, Lubomir and Huang, Thomas S},
  booktitle={European conference on computer vision},
  pages={679--692},
  year={2012},
  organization={Springer}
}

@inproceedings{zhu2012face,
  title={Face detection, pose estimation, and landmark localization in the wild},
  author={Zhu, Xiangxin and Ramanan, Deva},
  booktitle={2012 IEEE conference on computer vision and pattern recognition},
  pages={2879--2886},
  year={2012},
  organization={IEEE}
}

@inproceedings{sagonas2013300,
  title={300 faces in-the-wild challenge: The first facial landmark localization challenge},
  author={Sagonas, Christos and Tzimiropoulos, Georgios and Zafeiriou, Stefanos and Pantic, Maja},
  booktitle={Proceedings of the IEEE international conference on computer vision workshops},
  pages={397--403},
  year={2013}
}

@article{belhumeur2013localizing,
  title={Localizing parts of faces using a consensus of exemplars},
  author={Belhumeur, Peter N and Jacobs, David W and Kriegman, David J and Kumar, Neeraj},
  journal={IEEE transactions on pattern analysis and machine intelligence},
  volume={35},
  number={12},
  pages={2930--2940},
  year={2013},
  publisher={IEEE}
}

@inproceedings{koestinger2011annotated,
  title={Annotated facial landmarks in the wild: A large-scale, real-world database for facial landmark localization},
  author={Koestinger, Martin and Wohlhart, Paul and Roth, Peter M and Bischof, Horst},
  booktitle={2011 IEEE international conference on computer vision workshops (ICCV workshops)},
  pages={2144--2151},
  year={2011},
  organization={IEEE}
}

@article{wang2018mask,
  title={Mask-pose cascaded cnn for 2d hand pose estimation from single color image},
  author={Wang, Yangang and Peng, Cong and Liu, Yebin},
  journal={IEEE Transactions on Circuits and Systems for Video Technology},
  volume={29},
  number={11},
  pages={3258--3268},
  year={2018},
  publisher={IEEE}
}

@inproceedings{pavlakos2024reconstructing,
  title={Reconstructing hands in 3d with transformers},
  author={Pavlakos, Georgios and Shan, Dandan and Radosavovic, Ilija and Kanazawa, Angjoo and Fouhey, David and Malik, Jitendra},
  booktitle={Proceedings of the IEEE/CVF Conference on Computer Vision and Pattern Recognition},
  pages={9826--9836},
  year={2024}
}

@inproceedings{cao2019cross,
  title={Cross-domain adaptation for animal pose estimation},
  author={Cao, Jinkun and Tang, Hongyang and Fang, Hao-Shu and Shen, Xiaoyong and Lu, Cewu and Tai, Yu-Wing},
  booktitle={Proceedings of the IEEE/CVF International Conference on Computer Vision},
  pages={9498--9507},
  year={2019}
}

@article{banik2021novel,
  title={A Novel Dataset for Keypoint Detection of quadruped Animals from Images},
  author={Banik, Prianka and Li, Lin and Dong, Xishuang},
  journal={arXiv preprint arXiv:2108.13958},
  year={2021}
}

@techreport{WahCUB_200_2011,
	Title = {{The Caltech-UCSD Birds-200-2011 Dataset}},
	Author = {Wah, C. and Branson, S. and Welinder, P. and Perona, P. and Belongie, S.},
	Year = {2011},
	Institution = {California Institute of Technology},
	Number = {CNS-TR-2011-001}
}

@inproceedings{van2015building,
  title={Building a bird recognition app and large scale dataset with citizen scientists: The fine print in fine-grained dataset collection},
  author={Van Horn, Grant and Branson, Steve and Farrell, Ryan and Haber, Scott and Barry, Jessie and Ipeirotis, Panos and Perona, Pietro and Belongie, Serge},
  booktitle={Proceedings of the IEEE Conference on Computer Vision and Pattern Recognition},
  pages={595--604},
  year={2015}
}

@article{yu2021ap,
  title={Ap-10k: A benchmark for animal pose estimation in the wild},
  author={Yu, Hang and Xu, Yufei and Zhang, Jing and Zhao, Wei and Guan, Ziyu and Tao, Dacheng},
  journal={arXiv preprint arXiv:2108.12617},
  year={2021}
}

@article{yang2022apt,
  title={Apt-36k: A large-scale benchmark for animal pose estimation and tracking},
  author={Yang, Yuxiang and Yang, Junjie and Xu, Yufei and Zhang, Jing and Lan, Long and Tao, Dacheng},
  journal={Advances in Neural Information Processing Systems},
  volume={35},
  pages={17301--17313},
  year={2022}
}

@article{labuguen2021macaquepose,
  title={MacaquePose: a novel “in the wild” macaque monkey pose dataset for markerless motion capture},
  author={Labuguen, Rollyn and Matsumoto, Jumpei and Negrete, Salvador Blanco and Nishimaru, Hiroshi and Nishijo, Hisao and Takada, Masahiko and Go, Yasuhiro and Inoue, Ken-ichi and Shibata, Tomohiro},
  journal={Frontiers in behavioral neuroscience},
  volume={14},
  pages={581154},
  year={2021},
  publisher={Frontiers Media SA}
}

@article{li2019atrw,
  title={ATRW: a benchmark for Amur tiger re-identification in the wild},
  author={Li, Shuyuan and Li, Jianguo and Tang, Hanlin and Qian, Rui and Lin, Weiyao},
  journal={arXiv preprint arXiv:1906.05586},
  year={2019}
}

@inproceedings{joska2021acinoset,
  title={Acinoset: a 3d pose estimation dataset and baseline models for cheetahs in the wild},
  author={Joska, Daniel and Clark, Liam and Muramatsu, Naoya and Jericevich, Ricardo and Nicolls, Fred and Mathis, Alexander and Mathis, Mackenzie W and Patel, Amir},
  booktitle={2021 IEEE international conference on robotics and automation (ICRA)},
  pages={13901--13908},
  year={2021},
  organization={IEEE}
}

@inproceedings{ng2022animal,
  title={Animal kingdom: A large and diverse dataset for animal behavior understanding},
  author={Ng, Xun Long and Ong, Kian Eng and Zheng, Qichen and Ni, Yun and Yeo, Si Yong and Liu, Jun},
  booktitle={Proceedings of the IEEE/CVF conference on computer vision and pattern recognition},
  pages={19023--19034},
  year={2022}
}

@article{ye2022superanimal,
  title={SuperAnimal models pretrained for plug-and-play analysis of animal behavior},
  author={Ye, Shaokai and Filippova, Anastasiia and Lauer, Jessy and Vidal, Maxime and Schneider, Steffen and Qiu, Tian and Mathis, Alexander and Mathis, Mackenzie Weygandt},
  journal={arXiv preprint arXiv:2203.07436},
  volume={4},
  number={5},
  year={2022}
}

@article{pereira2019fast,
  title={Fast animal pose estimation using deep neural networks},
  author={Pereira, Talmo D and Aldarondo, Diego E and Willmore, Lindsay and Kislin, Mikhail and Wang, Samuel S-H and Murthy, Mala and Shaevitz, Joshua W},
  journal={Nature methods},
  volume={16},
  number={1},
  pages={117--125},
  year={2019},
  publisher={Nature Publishing Group US New York}
}

@article{graving2019deepposekit,
  title={DeepPoseKit, a software toolkit for fast and robust animal pose estimation using deep learning},
  author={Graving, Jacob M and Chae, Daniel and Naik, Hemal and Li, Liang and Koger, Benjamin and Costelloe, Blair R and Couzin, Iain D},
  journal={elife},
  volume={8},
  pages={e47994},
  year={2019},
  publisher={eLife Sciences Publications, Ltd}
}

@inproceedings{khan2020animalweb,
  title={Animalweb: A large-scale hierarchical dataset of annotated animal faces},
  author={Khan, Muhammad Haris and McDonagh, John and Khan, Salman and Shahabuddin, Muhammad and Arora, Aditya and Khan, Fahad Shahbaz and Shao, Ling and Tzimiropoulos, Georgios},
  booktitle={Proceedings of the IEEE/CVF conference on computer vision and pattern recognition},
  pages={6939--6948},
  year={2020}
}

@inproceedings{wu2016single,
  title={Single image 3d interpreter network},
  author={Wu, Jiajun and Xue, Tianfan and Lim, Joseph J and Tian, Yuandong and Tenenbaum, Joshua B and Torralba, Antonio and Freeman, William T},
  booktitle={European Conference on Computer Vision},
  pages={365--382},
  year={2016},
  organization={Springer}
}

@inproceedings{reddy2018carfusion,
  title={Carfusion: Combining point tracking and part detection for dynamic 3d reconstruction of vehicles},
  author={Reddy, N Dinesh and Vo, Minh and Narasimhan, Srinivasa G},
  booktitle={Proceedings of the IEEE conference on computer vision and pattern recognition},
  pages={1906--1915},
  year={2018}
}

@inproceedings{deepfashion2,
  title={Deepfashion2: A versatile benchmark for detection, pose estimation, segmentation and re-identification of clothing images},
  author={Ge, Yuying and Zhang, Ruimao and Wang, Xiaogang and Tang, Xiaoou and Luo, Ping},
  booktitle={Proceedings of the IEEE/CVF Conference on Computer Vision and Pattern Recognition},
  pages={5337--5345},
  year={2019}
}

@article{wang2016benchmark,
  title={A benchmark for comparison of dental radiography analysis algorithms},
  author={Wang, Ching-Wei and Huang, Cheng-Ta and Lee, Jia-Hong and Li, Chung-Hsing and Chang, Sheng-Wei and Siao, Ming-Jhih and Lai, Tat-Ming and Ibragimov, Bulat and Vrtovec, Toma{\v{z}} and Ronneberger, Olaf and others},
  journal={Medical image analysis},
  volume={31},
  pages={63--76},
  year={2016},
  publisher={Elsevier}
}

@article{joham2024implicit,
  title={Implicit Is Not Enough: Explicitly Enforcing Anatomical Priors inside Landmark Localization Models},
  author={Joham, Simon Johannes and Hadzic, Arnela and Urschler, Martin},
  journal={Bioengineering},
  volume={11},
  number={9},
  pages={932},
  year={2024},
  publisher={MDPI}
}

\clearpage

\title{GKDT: General Keypoint Detection Transformer (Supplementary Material)}

% ---------------------------------------------------------------
% TODO REVIEW: If the paper title is too long for the running head, you can set
% an abbreviated paper title here. If not, comment out.
\titlerunning{GKDT: General Keypoint Detection Transformer}

% TODO FINAL: Replace with your author list. 
% Include the authors' OCRID for the camera-ready version, if at all possible.
\author{Changsheng Lu\inst{1} \and
Yuxin Chen\inst{1} \and
Haokun Gui\inst{1} \and
Rong Wang\inst{2} \and
Jie Yang\inst{3} \and
Harry Yang\inst{1} \and
Anton van den Hengel\inst{4} \and
Jiaya Jia\inst{1}
}

% TODO FINAL: Replace with an abbreviated list of authors.
\authorrunning{C. Lu et al.}
% First names are abbreviated in the running head.
% If there are more than two authors, 'et al.' is used.

% TODO FINAL: Replace with your institution list.
\institute{Hong Kong University of Science and Technology, Hong Kong, China\\
\email{changshengluu@gmail.com}, \email{jia@cse.ust.hk} \and
Australian National University, Canberra, Australia \and 
Tencent Inc., China \and
Adelaide University, Adelaide, Australia\\
\email{anton.vandenhengel@adelaide.edu.au}}
% ---------------------------------------------------------------

\maketitle

\setcounter{table}{9}
\setcounter{equation}{3}
\setcounter{figure}{6}

\appendix

% if the hyperlink goes to the wrong place, we can add
% \usepackage[hypertexnames=false]{hyperref} option
Summary of this supplementary material is as follows:
\begin{itemize}%[label=$-$]
  \item \textbf{\S\ref{sec:appx:dynamic_imp_sampling}} provides further descriptions on \textbf{Dynamic Importance Sampling}, including the visualization of distributions during simulated sampling in Fig.~\ref{fig:distributions_plot}, and Algorithm~\ref{alg:dynamic-importance}.
  \item \textbf{\S\ref{sec:appx:full_dataset_splits}} gives \textbf{Full Dataset Splits for MegaKPT}.
  \item \textbf{\S\ref{sec:appx:add_exp_details}} provides \textbf{Additional Implementation Details}.
  \item \textbf{\S\ref{sec:appx:more_analysis}} presents \textbf{More Analysis}, such as study on the size of generated kernels. 
  \item \textbf{\S\ref{sec:appx:more_visualizations}} gives \textbf{More Visualizations} using our single GKDT model, including i) detection on more diverse object categories (Fig.~\ref{fig:vis_more_detections_gkd}) and ii) detection on in-the-wild videos (Fig.~\ref{fig:vis-video}).
\end{itemize}

% \textcolor{blue}{\emph{We will release the dataset, models, and codes. We hope our work can serve as a solid baseline in general keypoint detection for our vision community.}}

\phantomsection
\section{Dynamic Importance Sampling}\label{sec:appx:dynamic_imp_sampling}
\begin{figure*}[!tb]
  % \vspace{-0.2cm}
  % \setlength{\abovecaptionskip}{0.1cm}
  % \setlength{\belowcaptionskip}{-0.cm}
  \centering
  \includegraphics[width=\linewidth]{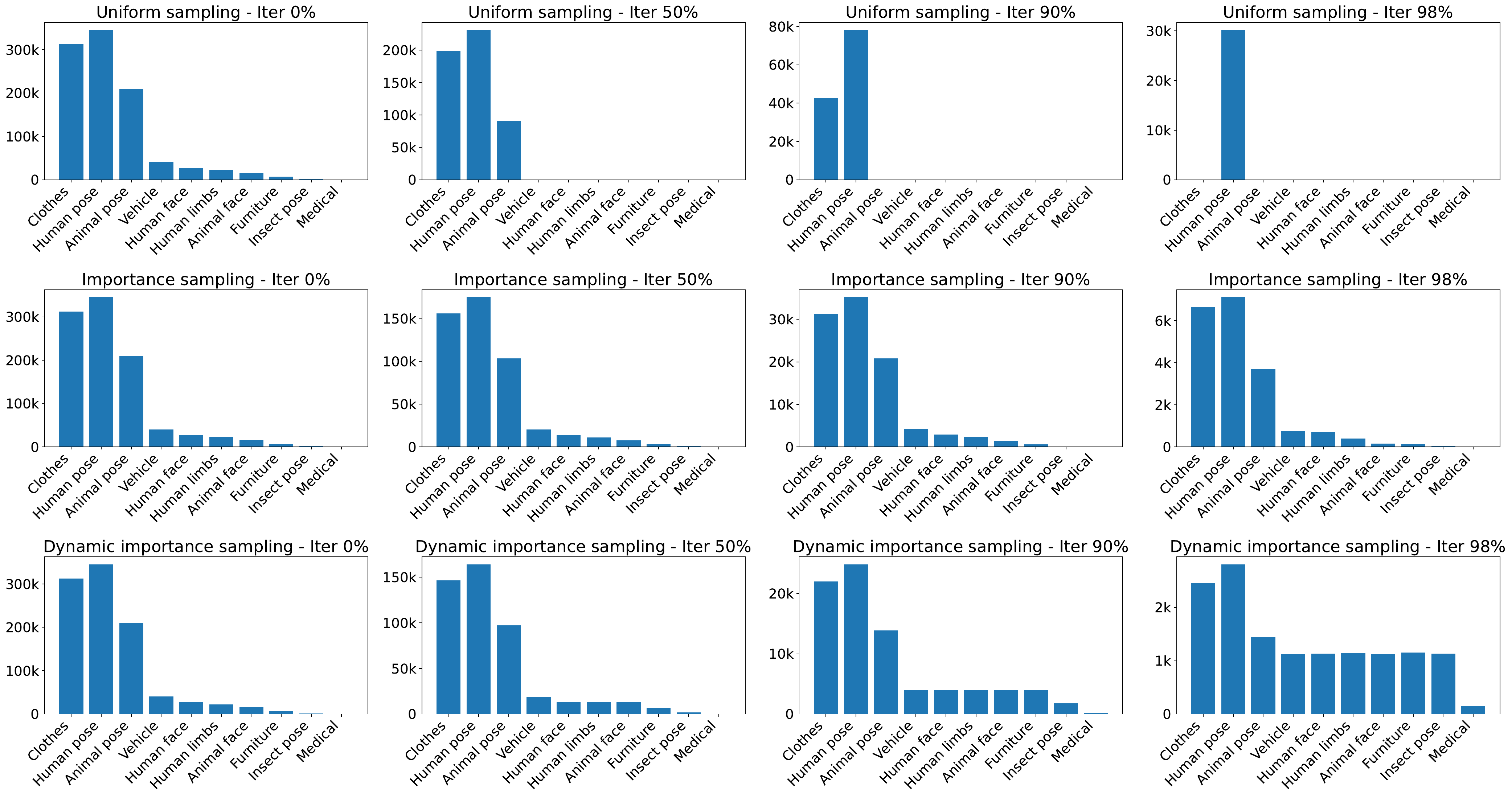}
  \caption{Simulation of three sampling strategies. The three rows show the distributions of \textit{uniform sampling}, \textit{importance sampling}, and our \textit{dynamic importance sampling} at iteration progresses 0\%, 50\%, 90\% and 98\%, respectively. Our dynamic importance sampling can adaptively adjust the distribution and mitigate the head-tail data imbalance. %especially when approaching the end of each progress.
  }
  \label{fig:distributions_plot}
  % \vspace{-0.1cm}
\end{figure*}

% We visualize the distributions in simulated sampling in Fig.~\ref{fig:distributions_plot}, and detail the dynamic importance sampling in Algorithm~\ref{alg:dynamic-importance}. Waiting to write... 
To understand how data distribution changes via our dynamic importance sampling, we perform simulation of data sampling using MegaKPT training set. The visualization result is shown in the third row of Fig.~\ref{fig:distributions_plot}. Compared to uniform sampling (1st row, Fig.~\ref{fig:distributions_plot}) and importance sampling (2nd row, Fig.~\ref{fig:distributions_plot}) which maintains strong head-tail data imbalance throughout iteration progress, after using our dynamic importance sampling (3rd row, Fig.~\ref{fig:distributions_plot}), the distribution against supercategories is gradually balanced, verifying its effectiveness in handling data imbalance. With the mechanism of dynamically judging if a sampled category from head classes (\ie, top $\gamma S$ classes), we perform data removal or not accordingly. In this way, in the early stage of data sampling, the sampled episodes dominantly come from head classes, but also have the opportunity from tail ones. The number of instances of those head classes will continuously reduce. Once one of them has fewer number of instances than any tail one, then it will be kicked out from top ranking classes and a tail one is in. Thus, in the late stage of data sampling, head-tail data imbalance is mitigated and there is larger opportunity for both head and tail classes to be sampled and added for model training. Note that the head categories are dynamically determined based on their remaining number of instances, which guarantees the priority of data removal from heads and the convergence of data balance. The step-by-step illustration of dynamic importance sampling is shown in Algorithm~\ref{alg:dynamic-importance}.

% Data distribution imbalance widely occurs in datasets, which leads a dilema for model training, especially in tail classes. 

% In this way, the number of samples from the head classes will decrease along with the progress of data sampling, and eventually reaching data balance with tail classes. Afterwards, the data samples from the tail classes will be drawn and removed from data pool, too. In this way, in the early stage of data sampling, most of samples are from head classes, but the samples from tail classes have the opportunity to be trained, too. In the late stage, both head and tail classes will have similar opportunity to be added to model training. Such a simple idea directly help us to greatly mitigate data imbalance issue, as evidenced by our experiments.

\begin{algorithm}[!tb]
\caption{Dynamic Importance Sampling}
\label{alg:dynamic-importance}
\Input{Data pool $\mathcal{D}$; Top percentile $\gamma \in (0,1]$}
\Output{Sampled episodes $\mathcal{E}=\{\mathbf{e}_i\}$}
\Initialize{$\mathcal{E} \gets \emptyset$; Partition $\mathcal{D}$ as $\{\mathcal{D}_i\}_{i=1}^S$ by $S$ super-categories, each with samples $|\mathcal{D}_i|$}
% \BlankLine
\While{True}{
    $P \gets \{p_i\}$, $p_i \gets \frac{|\mathcal{D}_i|}{\sum_{i} |\mathcal{D}_i|}$\;
    $c \gets \text{RandChoice}(1, S, \text{prob}=P)$      \textcolor{blue}{\hfill $\rhd$ Sample a super-category $c$ given P} \\ 
    $k \gets \gamma S$;                                   \textcolor{blue}{\hfill $\rhd$ Compute the number of head classes $k$} \\  
    \uIf{IsTopK($c$, $k$, $\{\mathcal{D}_i\}_{i=1}^S$)}{
      $\mathbf{e} \gets \text{SampleData}(\mathcal{D}_c)$\;
      $\mathcal{D}_c \gets \mathcal{D}_c \backslash \mathbf{e}$; \textcolor{blue}{\hfill $\rhd$ Sampled data removal if $c$ is in head classes}\\  
    }
    \Else{
      $\mathbf{e} \gets \text{SampleData}(\mathcal{D}_c)$; \textcolor{blue}{\hfill $\rhd$ Sampling an episode data but no removal}\\
    }
    $\mathcal{E} \gets \mathcal{E} \cup \{\mathbf{e}\}$\;
    \If{$|\mathcal{D}_c| < |\mathbf{e}|$}{
      $\mathcal{D} \gets \mathcal{D} \backslash \mathcal{D}_c$, $S \gets S-1$; \textcolor{blue}{$\qquad\qquad\quad$$\rhd$ Update data pool and categories}\\ 
    }
    \If{$\mathcal{D} = \emptyset$}{
      % \textbf{break}\;
      \Return $\mathcal{E}$. \\
    }
}
% \Return $\mathcal{E}=\{\mathbf{e}_i\}$\;
\end{algorithm}

\phantomsection
\section{Full Dataset Splits for MegaKPT}\label{sec:appx:full_dataset_splits}
The full splits are as follows:

\begin{itemize}
  \item \textbf{COCO \cite{lin2014microsoft}:} We follow the existing official splits where the training and validation sets have 262,465 and 11,004 annotated human pose instances, respectively.
  \item \textbf{Human-Art \cite{ju2023human}:} We follow the official training and validation splits for human instances.
  \item \textbf{300W \cite{sagonas2016300}:} We follow the standard training, validation, and test splits.
  \item \textbf{OneHand10K \cite{wang2018mask}:} We follow the official training and validation splits.
  \item \textbf{HInt \cite{pavlakos2024reconstructing}:} We combine all subsets within the training, validation and test, resulting in 12446, 1175, 3660 human instances, respectively.
  \item \textbf{Animal pose dataset \cite{cao2019cross}:} The entire dataset is for testing.
  \item \textbf{AwA pose \cite{banik2021novel}:} Following the previous few-shot keypoint detection benchmark \cite{lu2024openkd}, the AwA is split into 25 disjoint species for training and 10 for testing.
  \item \textbf{CUB \cite{WahCUB_200_2011}:} 100 species are used for training, 50 for validation, and 50 for testing \cite{lu2024openkd}.
  \item \textbf{NABird \cite{van2015building}:} 333 species are used for training, 111 for validation, and 111 for testing \cite{lu2024openkd}.
  \item \textbf{AP-10K \cite{yu2021ap}:} This dataset consists of 10,015 images from 54 animal species across 23 families. Each instance is annotated with 17 keypoints. We used 42 species for training and the remaining 12 species for testing.
  \item \textbf{MacaquePose \cite{labuguen2021macaquepose}:} We randomly split the macaque instances into training and test sets at the ratio of 0.8 by 0.2.
  \item \textbf{ATRW (tiger) \cite{li2019atrw}:} The official training and test splits are used.
  \item \textbf{Animal Kingdom \cite{ng2022animal}:} The official training and test splits are used.
  \item \textbf{TopViewMouse-5K \cite{ye2022superanimal}:} The official training and test splits are used.
  \item \textbf{Vinegar Fly \cite{pereira2019fast}:} We randomly split fly instances into training and test sets at the ratio of 0.8 by 0.2.
  \item \textbf{Desert Locust \cite{graving2019deepposekit}:} The same to Vinegar Fly \cite{pereira2019fast}.
  \item \textbf{AnimalWeb \cite{khan2020animalweb}:} This dataset has 22451 animal face images from 350 different animal categories and 21 orders. Similar to AwA, CUB, and NABird, we split them into disjoint categories of 250 as seen categories for training, while 100 as unseen ones for testing.
  \item \textbf{Keypoint-5 \cite{wu2016single}:} This furniture dataset includes 5 categories, which are bed, chair, sofa, swivelchair, and table. We follow standard splits for training and testing.
  \item \textbf{CarFusion \cite{reddy2018carfusion}:} This vehicle dataset has 3 categories, which are car, suv, and truck. We follow standard splits for training and testing.
  \item \textbf{DeepFashion2 \cite{deepfashion2}:} This is a large-scale clothes dataset that has 13 categories with 8 to 39 landmarks. We also follow official training and validation splits.
  \item \textbf{Cephalometric \cite{wang2016benchmark}:} This dataset has 400 medical scans of human skull for cephalometrics. We follow official training and test splits, which have 150 and 250 images, respectively.
  \item \textbf{Hand X-ray \cite{joham2024implicit}:} The entire dataset is for testing to test the generalization of our general keypoint detection model.
\end{itemize} 
The other datasets such as AFLW \cite{koestinger2011annotated}, APT-36K \cite{yang2022apt}, and AcinoSet (cheetah) \cite{joska2021acinoset} are only used for training. To ease the usage of this dataset, we release above splits at \url{https://github.com/AlanLuSun/General-Keypoint-Detection}.

% Following \cite{lin2014microsoft}, the split COCO training and validation sets have 262,465 and 11,004 annotated human pose instances, respectively. %The COCO validation set is used for testing.
% 300W \cite{sagonas2016300} is a human facial landmark dataset that combines other small sets, including HELLEN \cite{le2012interactive}, AFW \cite{zhu2012face}, IBUG \cite{sagonas2013300}, and LFPW \cite{belhumeur2013localizing}. We follow the standard training, validation, and test splits \cite{sagonas2016300}. For the OneHand10K hand pose dataset \cite{wang2018mask}, we also follow its standard training and validation splits.

\phantomsection
\section{Additional Implementation Details}\label{sec:appx:add_exp_details}
In our detection head, the upsampler $\mathcal{U}$ can be either bilinear upsampler or two deconv blocks, which are for upscaling the resolution of query image feature maps. We find that the bilinear upsampler with upscale ratio $u=4$ already performs well. All GKDT models are trained using learning rate of 1e-4 with Adam optimizer. 
% We also explore the upsampler with bilinear interpolation and replace bilinear with Deconv as ViTPose \cite{xu2023vitpose++}, the results show that bilinear upsampler is enough for our GKDT model. 

For general keypoint detection in single-object scenario, the detected keypoints of the object bounding box with the largest area/score are used in X-Pose \cite{yang2024x}, which we find that it yields best performance. For multi-object scenario, the detected bounding boxes of compared methods are filtered via a score threshold that gives best F-measure. This score threshold is automatically searched from 0 to 1 with increment of 0.05 for each dataset. A predicted object box is regarded as correct if its IoU to GT is larger or equal to 0.5. In this way, the false positive bounding boxes are eliminated and the detection gives highest quality, yielding a convincing comparison for subsequent keypoint detection.

\begin{table*}[!b]
  % \vspace{-0.3cm}
  \centering
  \caption{Study on kernel size. The testing scores of PCK@0.1 are reported.
  }
  \label{tab:study_kernel_reso}
  % \resizebox{\linewidth}{!}{  % resize table
    % \scriptsize %\footnotesize %\small
    % \fontsize{7}{8}\selectfont
    \setlength{\tabcolsep}{3.5pt}
    \begin{tabular}{lccccccccc}
     \toprule[1pt]
           \multirow{2}*{Variants} &\multicolumn{3}{c}{Animal pose} &\multicolumn{3}{c}{CUB} &\multicolumn{3}{c}{HInt} \\\cmidrule(lr){2-4}\cmidrule(lr){5-7}\cmidrule(lr){8-10}
                           &visual&text  &both  &visual&text  &both  &visual&text  &both  \\ \midrule[1pt]
     kernel $1 \times 1$   &81.11 &73.27 &76.87 &97.71 &98.58 &98.48 &69.79 &82.64 &81.13 \\
     kernel $3 \times 3$   &81.52 &73.89 &77.28 &97.94 &98.79 &98.70 &69.24 &82.15 &80.87 \\
     kernel $5 \times 5$   &81.67 &70.56 &75.83 &97.88 &98.73 &98.64 &69.91 &82.69 &81.59 \\
    \bottomrule[1pt]
    \end{tabular}
%  }  % resizebox
% \vspace{-0.3cm}
\end{table*}

\begin{figure*}[!tb]
  % \vspace{-0.2cm}
  % \setlength{\abovecaptionskip}{0.1cm}
  % \setlength{\belowcaptionskip}{-0.cm}
  \centering
  \includegraphics[width=\linewidth]{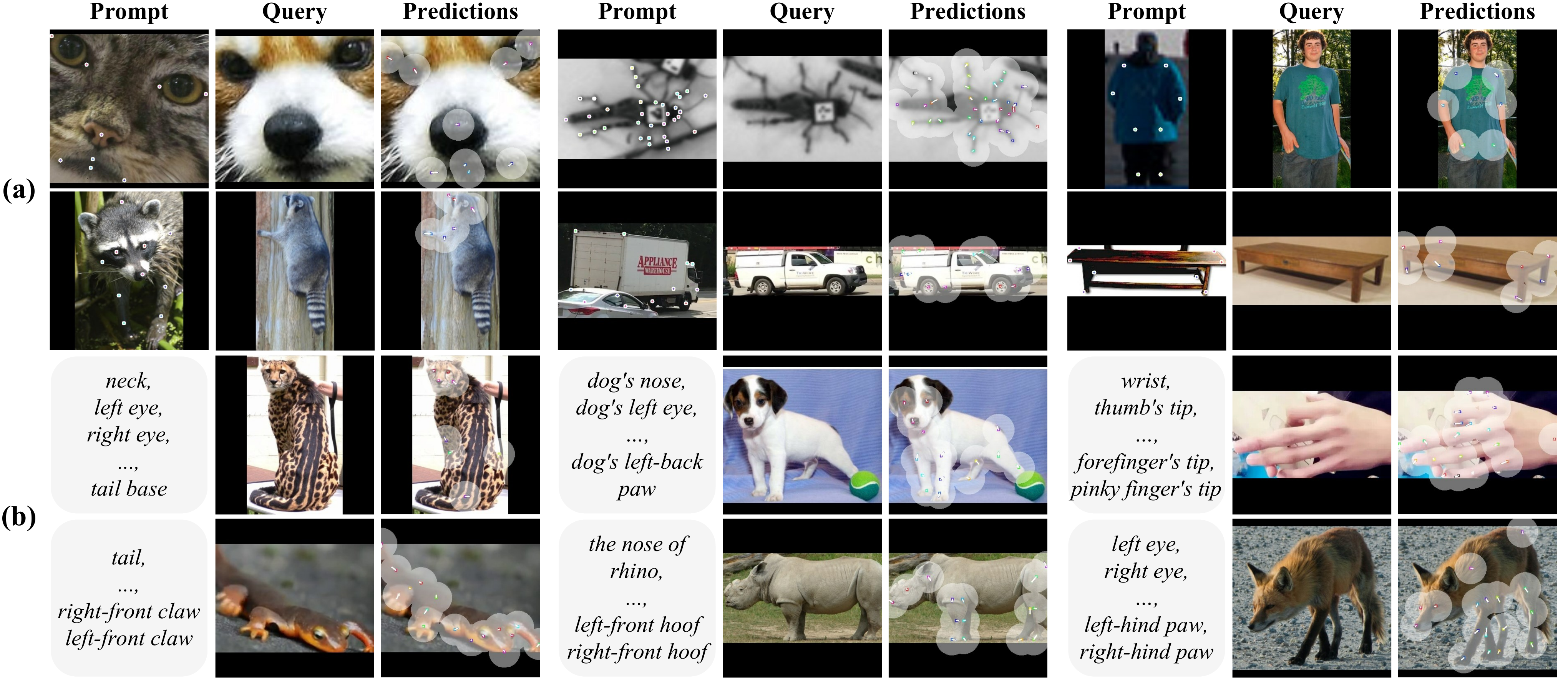}
  \caption{Visualization on more diverse categories, such as human and animal poses, faces, hands, insect poses, vehicles, and furniture. Our GKDT model can recognize keypoints on diverse objects, showing the excellent generality and performance.
  }
  \label{fig:vis_more_detections_gkd}
\end{figure*}

\begin{figure*}[!h]
  % \vspace{-0.2cm}
  % \setlength{\abovecaptionskip}{0.1cm}
  % \setlength{\belowcaptionskip}{-0.cm}
  \centering
  \includegraphics[width=\linewidth]{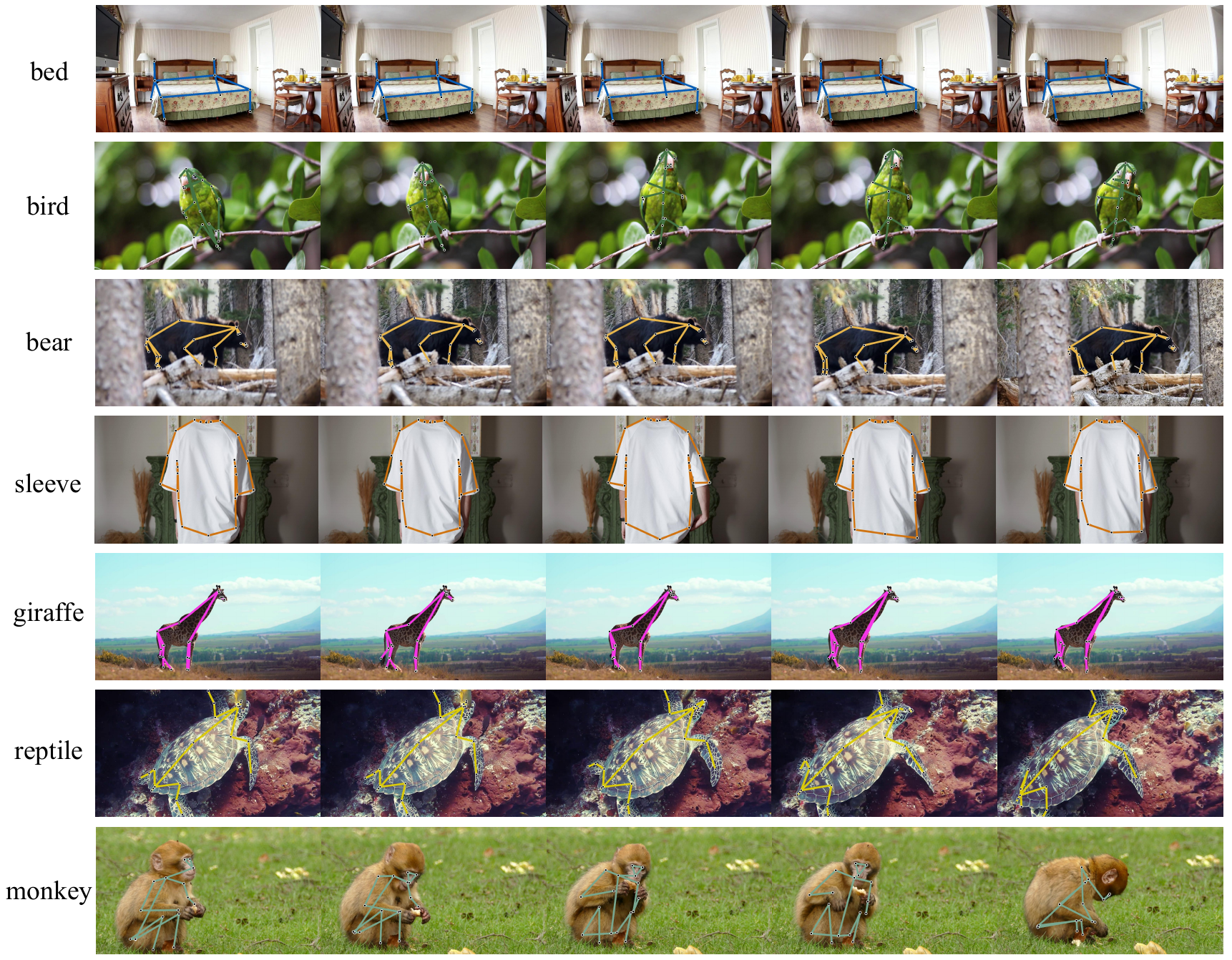}
  \caption{Visualization of zero-shot keypoint detection on in-the-wild videos. We supply the keypoint texts for each video clip to detect the corresponding keypoints. All results are obtained via a \emph{single GKDT model}.
  %Visualization on more diverse categories such as animals poses and faces, hand keypoints, insects, and vehicles. As one can see, our GKDT model can strongly recognize diverse keypoint on various objects, showing the excellent generality and performance.
  }
  \label{fig:vis-video}
  % \vspace{-0.4cm}
\end{figure*}

\phantomsection
\section{More Analysis}\label{sec:appx:more_analysis}
\noindent\textbf{Impacts of kernel size:} The size of generated kernels affects the perception field and the window to perform convolution to produce heatmap output. To study its impact, we let the generated kernels be with the sizes $1 \times 1$, $3 \times 3$, and $5 \times 5$, respectively. Table~\ref{tab:study_kernel_reso} shows that kernel size $s$ at $1 \times 1$ and $3 \times 3$ performs well overall, but may have performance degradation at $5 \times 5$, \eg, 70.56\% under text prompt in Animal pose dataset. We suspect that too large kernel size may introduce more unrelated context, thus yielding inferior keypoint heatmaps.

\phantomsection
\section{More Visualizations}\label{sec:appx:more_visualizations}
Firstly, beyond Fig.~\ref{fig:vis_heatmaps_gkd} and Fig.~\ref{fig:vis_multi_object_gkd} in main paper, we visualize general keypoint detection on more diverse object categories. As shown in Fig.~\ref{fig:vis_more_detections_gkd}, our GKDT model is capable to handle diverse objects under visual and text prompt.

Secondly, we wonder what the performance of our GKDT model will be on in-the-wild videos. To answer this question, we download several video clips from internet that cover various object categories, and evaluate our GKDT model with zero-shot keypoint detection. The qualitative results are shown in Fig.~\ref{fig:vis-video}. As one can see, our GKDT model achieves strong generality and performance on in-the-wild objects in time series of video frames, such as rigid bed, non-rigid clothes, and articulated animals, showcasing its practical applications in the real-world videos.
%and maintain good repeatability of keypoints in time series of video frames, 

\end{document}